\documentclass[letterpaper]{article} 
\usepackage{aaai25}  
\usepackage{times}  
\usepackage{helvet}  
\usepackage{courier}  
\usepackage[hyphens]{url}  
\usepackage{graphicx} 
\usepackage{natbib}  
\usepackage{caption} 
\usepackage{algorithm}
\usepackage{algorithmic}

\usepackage{amsmath}
\usepackage{amsfonts}
\usepackage{booktabs} 
\usepackage{multicol}
\usepackage{multirow}
\usepackage{placeins}
\usepackage{subcaption}
\usepackage[table]{xcolor}
\usepackage{hyperref}
\hypersetup{
	colorlinks=true,
	filecolor=red,      
	urlcolor=blue,
	citecolor=cyan,
}

\newcommand*{\pcr}{\fontfamily{pcr}\selectfont}

\usepackage{newfloat}
\usepackage{listings}
\DeclareCaptionStyle{ruled}{labelfont=normalfont,labelsep=colon,strut=off} 
\floatstyle{ruled}
\newfloat{listing}{tb}{lst}{}
\floatname{listing}{Listing}
\title{SparX: A Sparse Cross-Layer Connection Mechanism for\\Hierarchical Vision Mamba and Transformer Networks}

\author{
    Meng Lou,
    Yunxiang Fu,
    Yizhou Yu
}
\affiliations{
    School of Computing and Data Science, The University of Hong Kong \\
    \tt \small loumeng@connect.hku.hk, \tt \small yunxiang@connect.hku.hk, \tt \small yizhouy@acm.org

}

\usepackage{bibentry}

\begin{document}

\maketitle

\begin{abstract}
Due to the capability of dynamic state space models (SSMs) in capturing long-range dependencies with linear-time computational complexity, Mamba has shown notable performance in NLP tasks. This has inspired the rapid development of Mamba-based vision models, resulting in promising results in visual recognition tasks. However, such models are not capable of distilling features across layers through feature aggregation, interaction, and selection. Moreover, existing cross-layer feature aggregation methods designed for CNNs or ViTs are not practical in Mamba-based models due to high computational costs. Therefore, this paper aims to introduce an efficient cross-layer feature aggregation mechanism for vision backbone networks. Inspired by the Retinal Ganglion Cells (RGCs) in the human visual system, we propose a new sparse cross-layer connection mechanism termed SparX to effectively improve cross-layer feature interaction and reuse. Specifically, we build two different types of network layers: ganglion layers and normal layers. The former has higher connectivity and complexity, enabling multi-layer feature aggregation and interaction in an input-dependent manner. In contrast, the latter has lower connectivity and complexity. By interleaving these two types of layers, we design a new family of vision backbone networks with sparsely cross-connected layers, achieving an excellent trade-off among model size, computational cost, memory cost, and accuracy in comparison to its counterparts. For instance, with fewer parameters, SparX-Mamba-T improves the top-1 accuracy of VMamba-T from 82.5\% to 83.5\%, while SparX-Swin-T achieves a 1.3\% increase in top-1 accuracy compared to Swin-T. Extensive experimental results demonstrate that our new connection mechanism possesses both superior performance and generalization capabilities on various vision tasks. Code is publicly available at \url{https://github.com/LMMMEng/SparX}.
\end{abstract}

\par
\begin{figure}[t]
    \centering
\includegraphics[width=0.4675\textwidth]{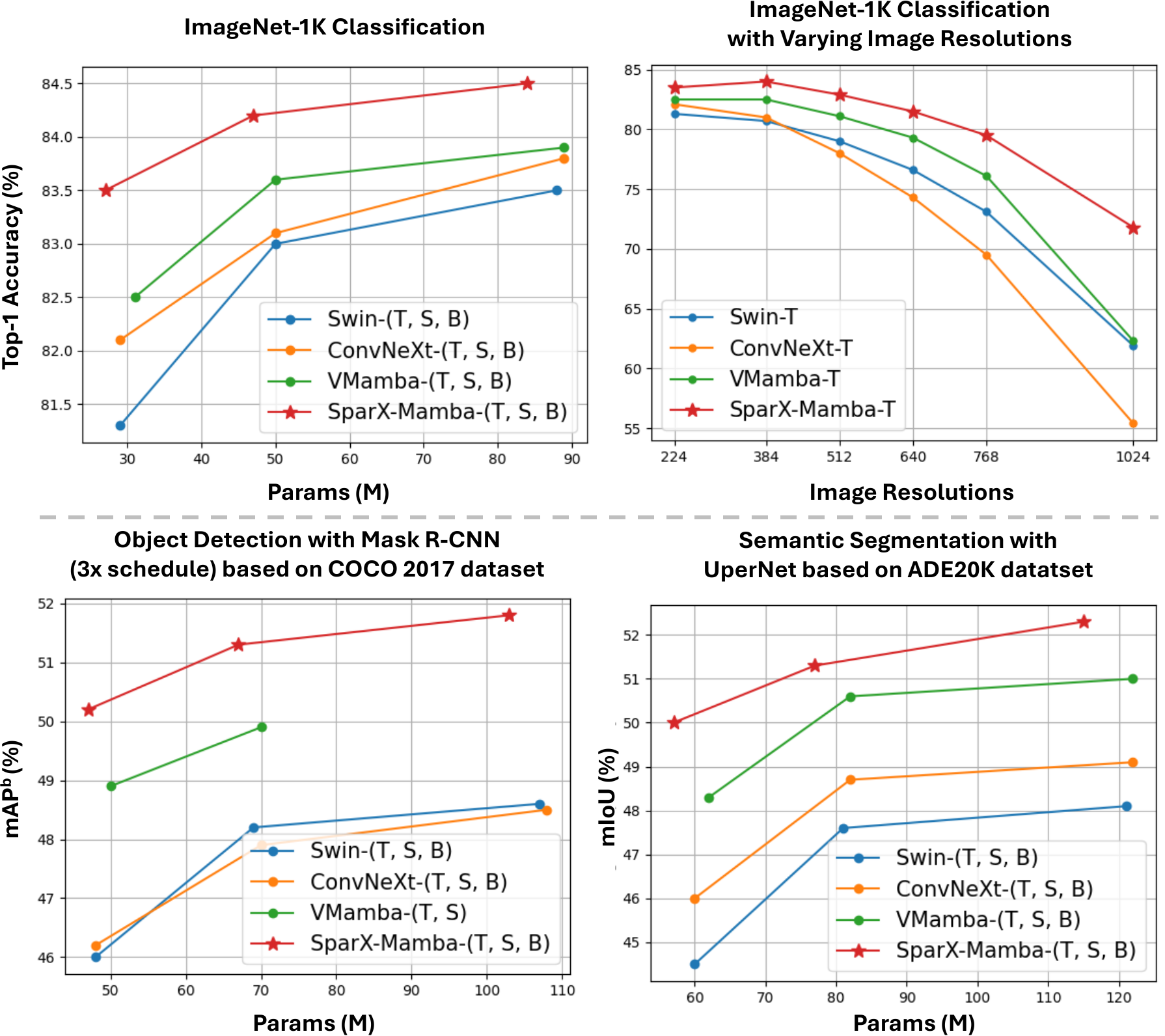}
    \caption{Performance comparison between SparX-Mamba and other methods on various vision tasks.}
    \label{intro_acc}
\end{figure}

\section{Introduction}
Contextual modeling plays a vital role in computer vision tasks, including image classification, object detection, and semantic segmentation. Recent advancements have explored large-kernel convolutions to enlarge receptive fields~\cite{liu2022convnet, ding2022scaling, liu2022more, ding2023unireplknet} and Vision Transformers (ViTs)~\cite{dosovitskiy2020image,touvron2021training,liu2021swin, wang2021pyramid} to possess powerful long-range modeling capability. However, both CNNs with large convolution kernels and Vision Transformers exhibit limitations. The performance of CNNs still lags behind advanced ViTs while the latter suffers from substantial computational costs when processing high-resolution inputs, due to its quadratic complexity.
\par
Empowered by dynamic State Space Models (SSMs) capable of modeling long-range dependencies with near-linear complexity, Mamba~\cite{gu2023mamba} has demonstrated promising performance in natural language processing (NLP) tasks. Building upon this success, researchers have proposed Mamba-based models for vision tasks~\cite{zhu2024vision,yang2024plainmamba,liu2024vmamba,huang2024localmamba,behrouz2024mambamixer,pei2024efficientvmamba}. Such work has made successful attempts to introduce linear-time long-range modeling capability into vision models by leveraging SSMs, which are primarily used to capture spatial contextual information within the same layer. Nonetheless, there might exist both complementarity and redundancy among token features in different layers as they capture image characteristics and semantics at different levels and granularities. Thus features across different layers need to be distilled to extract useful information and remove redundancy through aggregation, interaction, and selection. However, existing Mamba-based vision models do not possess such an ability to conduct feature distillation across layers, hindering the realization of the full power of SSMs.
\par
Although cross-layer feature interaction and reuse have been shown to effectively improve performance in both CNNs and Vision Transformers, such as DenseNet~\cite{huang2017densely} and FcaFormer~\cite{zhang2023fcaformer}, these methods are not directly applicable to Mamba-based vision models. As shown in Table~\ref{tab:compare_densenet}, when VMamba is restructured as a DenseNet-style Network (DSN), its throughput drops by almost 50\%, accompanied by a 1GB increase in GPU memory usage. Meanwhile, as shown in Table~\ref{tab:compare_dmca}, using Spatial Cross-Attention (SCA) in FcaFormer as a multi-layer feature interaction module results in a more than 80\% increase in GPU memory usage compared to simple feature concatenation, due to the quadratic complexity of attention calculations. Therefore, designing an efficient connectivity pattern to facilitate cross-layer feature interaction in Mamba-based models poses a significant challenge.
\par
Unlike existing works that primarily focus on improving Mamba-based token mixers, in this work, we propose a novel architectural design that significantly boosts the performance of SSM-based vision models. Our design draws inspiration from Retinal Ganglion Cells (RGCs) in the human visual system. RGCs serve as the information transmission hub in the human visual system, bridging the retinal input and the visual processing centers in the central nervous system~\cite{kim2021retinal}. Due to the complex neural architecture and diverse cell composition within the RGC layer~\cite{curcio1990human, watson2014formula}, it has a larger number of connections with other cell layers in the visual system, thereby fostering intricate neural interactions with other layers. Inspired by this biological mechanism, we propose a novel sparse cross-layer connection mechanism named SparX. We define two types of SSM-based basic layers, namely, normal layers and ganglion layers. The former has less information flow, taking a single input from the preceding layer and connecting with a small number of subsequent layers. In contrast, the latter establishes a larger number of connections with subsequent layers and encodes inputs from multiple preceding layers. The combination of ganglion and normal layers mimics the combination of RGC layers and non-RGC layers within the human visual system.
\par
Existing literature~\cite{d2020retinal} also confirms that RGCs generate diverse intercellular communications including both RGC-RGC and RGC-non-RGC interactions. Inspired by this premise, we incorporate a new Dynamic Multi-layer Channel Aggregator (DMCA) into ganglion layers, aiming to efficiently facilitate adaptive feature aggregation and interaction across multiple layers. Furthermore, to maintain speed and lower memory consumption, we introduce a cross-layer sliding window, which only permits each ganglion layer to connect with other ganglion layers in the same sliding window.
\par
By hierarchically interleaving ganglion and normal layers in a feedforward network, we design a novel Mamba-based vision backbone, namely, Vision Mamba with sparsely cross-connected layers (SparX-Mamba). In addition to inherent advantages of Mamba-based models, our SparX-Mamba achieves an excellent trade-off between performance and computational cost. Notably, our SparX-Mamba differs from conventional vision models that are constructed by simply stacking the same layer design, resulting in similar feature extraction capabilities at every layer. In contrast, the two types of functionally distinct layers of SparX-Mamba facilitate a model to extract more diverse feature representations, thereby enhancing performance. It is worth noting that our SparX has also been successfully applied to hierarchical vision Transformers, such as Swin \cite{liu2021swin}, and the resulting models also have the aforementioned advantages over the original Transformer models. For example, our SparX-Swin-T achieves a 1.3\% improvement in top-1 accuracy compared to Swin-T.
\par
As shown in Figure~\ref{intro_acc}, the proposed SparX-Mamba demonstrates superior performance over recently proposed VMamba as well as classical CNN and Transformer-based models, with a smaller number of parameters (Params). For instance, when compared with VMamba-T, SparX-Mamba-T achieves a top-1 accuracy of 83.5\% on ImageNet-1K, remarkably surpassing the accuracy of VMamba-T (82.5\%). When integrated with UperNet for semantic segmentation on the ADE20K dataset, SparX-Mamba-T outperforms VMamba-T by 1.7\% in mIoU. Furthermore, when integrated into Mask R-CNN for object detection on the COCO 2017 dataset, SparX-Mamba-T exceeds VMamba-T by 1.3\% in AP$^{b}$ and even outperforms VMamba-S, whose complexity is nearly 1.5 times that of SparX-Mamba-T. Meanwhile, both the small and base versions of SparX-Mamba also exhibit notable performance improvements over VMamba.
\par
In summary, our main contributions are threefold: First, inspired by RGCs in the human visual system, we propose a novel skip-connection mechanism named SparX, which sparsely and dynamically configures cross-layer connections, thereby enabling diverse information flow and improved feature distillation. Second, on the basis of Mamba and Transformer, we propose two versatile vision backbones, SparX-Mamba and SparX-Swin, constructed from two types of functionally distinct layers, one of which is responsible for communications between relatively distant layers. Third, we conduct extensive experiments on image classification, object detection, and semantic segmentation tasks. Results demonstrate that our new network architecture achieves a remarkable trade-off between performance and computational cost.

\par\begin{figure*}[t]
    \centering
    \includegraphics[width=0.90\textwidth]{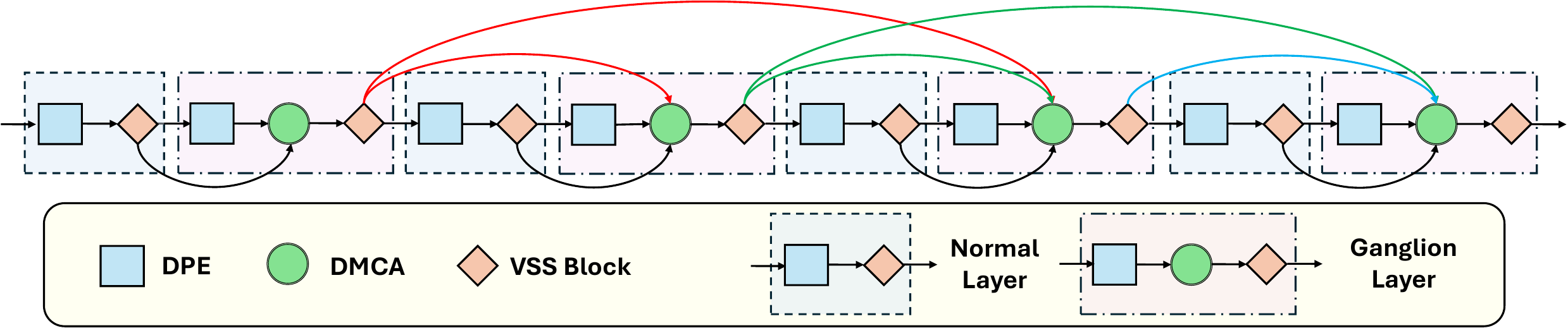} 
    \caption{A concrete example of proposed SparX with \textbf{S}=2 and \textbf{M}=2.\vspace{-0mm}}
    \label{SparX_fig}
\end{figure*}

\section{Related Work}
\textbf{CNNs}. Convolutional Neural Networks (CNNs) have emerged as the standard deep learning architecture in computer vision \cite{simonyan2014VGG,he2016deep, huang2017densely}. In the realm of modern CNNs, a notable shift has occurred from traditional small convolution kernels to a design focused on larger kernels. Noteworthy examples of this trend include ConvNeXt~\cite{liu2022convnet}, RepLKNet~\cite{ding2022scaling,ding2023unireplknet}, and SLaK~\cite{liu2022more}. Recently, InceptionNeXt~\cite{yu2023inceptionnext} has combined classical Inception networks with ConvNeXt, resulting in notable performance improvements.
\par
\textbf{Vision Transformers}. 
To adapt Transformers from NLP tasks to computer vision tasks, a vision Transformer (ViT)~\cite{dosovitskiy2020image} splits an image into visual tokens through patch embedding, thereby enabling multi-head self-attention (MHSA) to learn token-to-token dependencies. To further generate hierarchical feature representations and improve model efficiency, many subsequent works have adopted pyramidal architecture designs, including window attention~\cite{liu2021swin,dong2022cswin,pan2023slide}, sparse attention~\cite{yang2021focal,wang2022pvt,ren2022shunted,wu2022p2t}, and conv-attention hybrid models~\cite{dai2021coatnet,xiao2021early,li2022efficientformer,li2022uniformer,lou2023transxnet,metaformer2024}.
\par
\textbf{Vision Mamba}. Since Mamba~\cite{gu2023mamba} has achieved outstanding performance in NLP tasks, many researchers have transferred Mamba to computer vision tasks. As the core of Mamba, SSM can model long-range dependencies with near-linear complexity and has shown excellent performance in vision tasks. For instance, ViM~\cite{zhu2024vision} introduces a bidirectional SSM module and constructs an isotropic architecture like ViT~\cite{dosovitskiy2020image}. Likewise, PlainMamba~\cite{yang2024plainmamba} also builds an isotropic architecture with continuous 2D scanning. VMamba~\cite{liu2024vmamba} extends the scanning order to include four directions and is an early SSM-based hierarchical architecture.  Subsequently, a series of hierarchical vision Mamba models have been proposed, including MambaMixer~\cite{behrouz2024mambamixer}, LocalMamba~\cite{huang2024localmamba}, EfficientVMamba~\cite{pei2024efficientvmamba}, and MSVMamba \cite{shi2024multi}.
\par
\textbf{Skip-connections}. ResNet~\cite{he2016deep} introduces residual connections for deep CNNs, alleviating the issues of vanishing and exploding gradients with bypassing shortcuts. To diversify information flow, DenseNet~\cite{huang2017densely} further incorporates dense connections, which compute the input of a layer using the collection of outputs from all preceding layers. DPN~\cite{chen2017dual} makes use of the skip-connections in both ResNet and DenseNet to build a dual-pathway network. In the domain of dense prediction tasks, U-Net~\cite{ronneberger2015u} and FPN~\cite{lin2017feature} leverage skip-connections to bridge the gap between low-level details from the encoder and high-level contexts from the decoder. This work introduces a novel skip-connection mechanism that dynamically configures sparse cross-layer connections, resulting in improved performance in comparison to existing methods.
\par
\textbf{Cross-feature Attention}. Cross-feature attention can enhance feature representation by improving interactions among different features. For instance, many works \cite{chen2021crossvit,lee2022mpvit,wang2023crossformer++,ren2022shunted,wu2022p2t} proposed to perform cross-feature attention by generating multi-scale tokens, which are then fed into self-attention or cross-attention layers to model multi-scale interactions. Recently, FcaFormer \cite{zhang2023fcaformer} introduced cross-attention between the spatial tokens of a layer and representative tokens from previous layers to model cross-layer feature interactions. In contrast to existing methods, we introduce a novel channel-wise cross-attention mechanism that dynamically integrates feature channels from preceding layers, thereby facilitating efficient cross-layer feature interactions.

\section{Method}
\label{sec:Method}
\subsection{Sparse Cross-Layer Connections}
\label{sec:SparX}
\textbf{Overview.} Inspired by how Retinal Ganglion Cells (RGCs) function in the human visual system, we introduce a novel sparse cross-layer connection mechanism named SparX. The goal is efficiently modeling cross-layer communications, generating diverse information flows, and improving feature reuse within Mamba-based architectures. As shown in Figure~\ref{SparX_fig}, SparX has three building components: Dynamic Position Encoding (DPE)~\cite{chu2022conditional}, Mamba block, and the newly-introduced Dynamic Multi-layer Channel Aggregator (DMCA). Among them, DPE leverages a residual 3$\times$3 depthwise convolution (DWConv), which finds widespread adoption in modern vision backbone networks~\cite{li2022uniformer,chu2021twins,guo2022cmt}. The Mamba block effectively captures long-range dependencies with a near-linear computational complexity. However, the vanilla Mamba block~\cite{gu2023mamba} is unsuitable for being directly embedded into vision backbones. In this regard, we utilize the VSS block, which has shown promising performance since its introduction in VMamba~\cite{liu2024vmamba}. To enhance channel mixing, a ConvFFN \cite{wang2022pvt} is incorporated within VSS following \cite{shi2024multi}. The proposed DMCA dynamically aggregates, selects, and encodes cross-layer features, giving rise to powerful and robust feature representations. On top of these basic components, two types of network layers are constructed in SparX: ganglion layers (``{\pcr{DPE}} $\rightarrow$ {\pcr{DMCA}} $\rightarrow$ {\pcr{VSS Block}}") and normal layers (``{\pcr{DPE}} $\rightarrow$ {\pcr{VSS Block}}"). These two layers are interleaved in a feedforward network to enable effective modeling of cross-layer feature interactions in a sparse manner.
\par
\textbf{Connection Rules.} In a vanilla dense connection~\cite{huang2017densely}, each layer is connected to all preceding layers. Despite the relatively low FLOPs of DenseNet, the storage of many preceding feature maps, which need to be repeatedly accessed during a forward pass, gives rise to high memory cost and low speed. Although efforts~\cite{pleiss2017memory,huang2019convolutional} have been made to lower memory cost, dense connections still face challenges when it is necessary to scale models to deeper and wider architectures. Consequently, the vanilla dense connection mechanism is too computationally heavy to use in Mamba-based models whose token mixer is more complex than standard convolutions. To this end, we propose SparX, which can be efficiently integrated into recent Mamba-based architectures. In SparX, we introduce two new rules, including {sparse placement of ganglion layers} and using {a cross-layer sliding window} to control the density of cross-layer connections.

\begin{figure}[!b]
    \centering
    \includegraphics[width=0.495\textwidth]{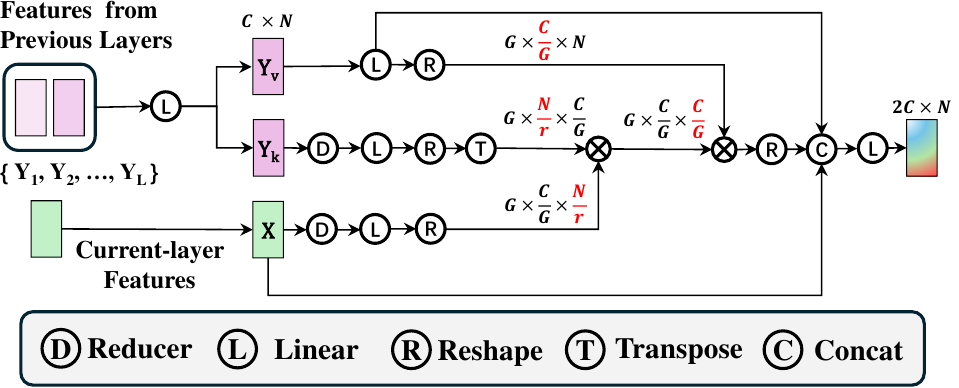} 
    \caption{An illustration of the proposed DMCA. The red font indicates the dimensions in matrix multiplications.}
    \label{mcda_fig}
\end{figure}

\textbf{Sparse Ganglion Layers} aim to designate a subset of evenly spaced layers as ganglion layers while all remaining layers are normal layers. To control the density of ganglion layers, we define a hyperparameter called stride (\textbf{S}), which is one plus the number of normal layers between two nearest ganglion layers. We further define two types of cross-layer connections: (1) Intra-connection between a ganglion layer and a normal layer; (2) Inter-connection between two ganglion layers. To set up sparse cross-layer connections, a ganglion layer only has intra-connections with the normal layers between the closest preceding ganglion layer and itself while it builds inter-connections with multiple preceding ganglion layers. The rationale behind such cross-layer connections is that a ganglion layer can be regarded as an information exchange hub, which gathers information from the closest normal layers and exchanges it with other ganglion layers. To ensure that the final output of a network or stage contains rich semantic information, the last layer of a network or stage is typically a ganglion layer. As an example, if we have an 8-layer network and set the stride (\textbf{S}) to 2, the indices of the normal layers are $\left \{ 1, 3, 5, 7 \right \}$ while the indices of the ganglion layers are $\left \{ 2, 4, 6, 8 \right \}$.
\par
\textbf{Cross-layer Sliding Window} is proposed to further improve computational efficiency, inspired by spatial sliding windows. The motivation behind this design is that, despite using the aforementioned sparse connections, deeper networks may still incur high memory costs due to the need to store and access a number of earlier feature maps. To this end, we introduce another hyperparameter \textbf{M}, which restricts a ganglion layer to have inter-connections with the \textbf{M} closest preceding ganglion layers only. Based on these two new rules, even without direct connections, information can still flow from shallower layers to deeper layers quickly through a relatively small number of intra-connections and inter-connections. Figure \ref{SparX_fig} illustrates an example of SparX with \textbf{S}=2 and \textbf{M}=2.
\par
\textbf{Dynamic Multi-layer Channel Aggregator (DMCA)}. To selectively retrieve complementary features from previous layers and dynamically model multi-layer interactions, we propose an efficient Dynamic Multi-layer Channel Aggregator (DMCA). As depicted in Figure~\ref{mcda_fig}, let $\mathbf{X} \in \mathbb{R}^{C\times N}$ represents the features of a ganglion layer, and $ \left \{\mathbf{Y_1, Y_2, \cdots, Y_L}  \right \}  \in \mathbb{R}^{C \times N}$ denote features from preceding ganglion and normal layers, where \textit{C} and \textit{N} represent the channel and spatial dimensions, respectively, while \textbf{L} denotes the total number of features from prior layers. First, we concatenate the features from preceding layers and project the channel dimension to 2\textit{C} using a linear layer. Then, we evenly split the projected features into two segments along the channel dimension, $\mathbf{Y_k}$ and $\mathbf{Y_v}$. Afterwards, we design a cross-layer grouped channel attention (CGCA) mechanism, where \textit{query}, \textit{key}, and \textit{value} refer to $\mathbf{X}$, $\mathbf{Y_k}$, and $\mathbf{Y_v}$, respectively. Specifically, we divide the channels into multiple groups, and channel-wise cross-attention is computed within each group. The number of groups (G) is always set to 4 in our experiments. Consequently, in our CGCA, the size of the overall attention map $\mathbf{A}$ for all groups is $G \times \frac{C}{G} \times \frac{C}{G}$, which avoids high computational and memory overhead when large spatial resolutions are used.
\par
On the other hand, channel-wise cross-attention computes an attention matrix that measures the similarity between every pair of transformed channels from \textit{query} and \textit{key}, and the spatial dimension becomes eliminated. Note that the spatial resolution directly affects the cost of computing this attention matrix because it determines the dimensionality of every channel from \textit{query} and \textit{key}. One key observation is that the size of the attention matrix is independent of the spatial dimensions of \textit{query} and \textit{key}, whose spatially reduced versions can be used for computing an approximation of the original attention matrix. Based on this, we employ a spatial reduction strategy to improve computational efficiency. Specifically, we use a spatial reducer to compress the number of spatial tokens in \textit{query} and \textit{key} from \textit{N} to \textit{N}/\textit{r}. In our experiments, the rule for setting \textit{r} is making \textit{N}/\textit{r} equal to the number of tokens in the final stage of the network, following PVT~\cite{wang2021pyramid, wang2022pvt}. Although this operation might appear similar to spatial reduction used in PVT, it serves different purposes. In PVT, spatial reduction is exploited to compute spatial token-to-region attention with the reduction applied to \textit{key} and \textit{value}. In our case, we employ spatial reduction to efficiently compute channel-to-channel attention with the reduction applied to \textit{query} and \textit{key}. Once the attention matrix has been computed, it is used to aggregate the channels of \textit{value}, which maintains the original spatial resolution. The final feature $\mathbf{Z}$ from channel-wise cross-attention can be regarded as the selected feature of \textit{value} under the guidance of \textit{query}. Finally, we concatenate $\mathbf{X}$, $\mathbf{Y_v}$, and $\mathbf{Z}$, followed by a linear layer to project the number of channels down to 2\textit{C}. Mathematically, the operations of DMCA are summarized below:
\begin{equation}
\centering
\begin{gathered}
\mathbf{Y_k},\mathbf{Y_v} = \mathrm{Split}(W_1(\mathrm{Cat}(\mathbf{Y_1},\mathbf{Y_1}, \cdots, \mathbf{Y_L})))
\\
\mathrm{Q}=W_2(R_1(\mathbf{X})), \mathrm{K}=W_3(R_2(\mathbf{Y_k})), \mathrm{V}=W_4(\mathbf{Y_v})
\\
\mathbf{Z} = \frac{\mathrm{{Softmax}} (\mathrm{Q}\mathrm{K} ^\mathrm{T} )}{\sqrt{N/r} } \mathrm{V}
\end{gathered} 
\end{equation}
where $W_i$ refers to a linear layer used for channel projection, i.e., $W_1 \in \mathbb{R}^{2C \times LC}$ and $\left\{W_2, W_3, W_4  \right\} \in \mathbb{R}^{C \times C}$. Meanwhile, $R_i$ denotes a spatial reducer implemented as a strided DWConv, while $\left\{ \mathrm{Q}, \mathrm{K} \right\} \in \mathbb{R}^{G \times \frac{C}{G} \times \frac{N}{r}}$ and $\mathrm{V} \in \mathbb{R}^{G \times \frac{C}{G} \times N}$.
\par
SparX introduces two flexible hyperparameters, namely \textbf{S} and \textbf{M}, making it easier and more efficient to extend the network deeper and wider. In addition, the incorporation of DMCA enables strong cross-layer feature interactions, giving rise to more powerful and robust representations.

\begin{table}[b]
  \centering
    \fontsize{9}{12}\selectfont 
    \setlength{\tabcolsep}{1mm}
    \begin{tabular}{ccccc}
    \toprule
    SparX-Mamba & Channels & Blocks & S, M\\
    \midrule
    Tiny  & [96, 192, 320, 512] & [2, 2, 7, 2] & [2, 3]  \\
    Small & [96, 192, 328, 544] & [2, 2, 17, 2] & [3, 3]  \\
    Base  & [120, 240, 396, 636] & [2, 2, 21, 3] & [3, 3]  \\
    \bottomrule
    \end{tabular}%
\caption{Detailed configurations of the three variants of SparX-Mamba.}
  \label{tab:model-variants}%
\end{table}

\begin{table}[!t]
    \centering
    \fontsize{9}{12}\selectfont 
    \setlength{\tabcolsep}{1mm}
    \begin{tabular}{lccc}
    \toprule
    Method & F (G) & P (M) & Acc. (\%) \\
    \midrule
    ConvNeXt-T & 4.5   & 29    & 82.1  \\
    InceptionNeXt-T & 4.2   & 28    & 82.3  \\
    SLaK-T & 5.0   & 30    & 82.5  \\
    UniRepLKNet-T & 4.9   & 31    & 83.2  \\
    MambaOut-T & 4.5   & 27    & 82.7  \\
    Focal-T & 4.9   & 29    & 82.2  \\
    PVTv2-B2 & 4.0   & 25    & 82.0  \\
    Swin-T & 4.5   & 29    & 81.3  \\
    CSWin-T & 4.5   & 23    & 82.7  \\
    UniFormer-S & 3.6   & 22    & 82.9  \\
    MambaOut-T & 4.5     & 27    & 82.7  \\
    ViM-S & -     & 26    & 81.6  \\
    PlainMamba-L2 & 8.1   & 25    & 81.6  \\
    ViM2-T & -     & 20    & 82.7  \\
    EfficientVMamba-B & 4.0   & 33    & 81.8  \\
    VMamba-T & 4.9   & 31    & 82.5  \\
    LocalVMamba-T & 5.7   & 26    & 82.7  \\
    MSVMamba-T & 4.6   & 33    & 82.8  \\
    \rowcolor{gray!20}\textbf{SparX-Mamba-T} & 5.2   & 27    & $\mathbf{83.5}$  \\
    \midrule
    ConvNeXt-S & 8.7   & 50    & 83.1  \\
    InceptionNeXt-S & 8.4   & 49    & 83.5  \\
    SLaK-S & 9.8   & 55    & 83.8  \\
    UniRepLKNet-S & 9.1   & 56    & 83.9  \\
    MambaOut-S & 9.0   & 48    & 84.1 \\
    Focal-S & 9.4   & 51    & 83.6  \\
    PVTv2-B4 & 10.1  & 63    & 83.6  \\
    Swin-S & 8.7   & 50    & 83.0  \\
    UniFormer-B & 8.3   & 50    & 83.9  \\
    PlainMamba-L3 & 14.4  & 50    & 82.3  \\
    ViM2-S & -     & 43    & 83.7  \\
    VMamba-S & 8.7   & 50    & 83.6  \\
    LocalVMamba-S & 11.4  & 50    & 83.7  \\
    \rowcolor{gray!20}\textbf{SparX-Mamba-S} & 9.3   & 47    & $\mathbf{84.2}$ \\
    \midrule
    ConvNeXt-B & 15.4  & 89    & 83.8  \\
    InceptionNeXt-B & 14.9  & 87    & 84.0  \\
    SLaK-B & 17.1  & 95    & 84.0  \\
    MambaOut-B & 15.8   & 85    & 84.2 \\
    Focal-B & 16.4  & 90    & 84.0  \\
    PVTv2-B5 & 11.8  & 82    & 83.8  \\
    Swin-B & 15.4  & 88    & 83.5  \\
    CSWin-B & 15.0  & 78    & 84.2  \\
    ViM2-B & -     & 74    & 83.9  \\
    VMamba-B & 15.4  & 89    & 83.9  \\
    \rowcolor{gray!20}\textbf{SparX-Mamba-B} & 15.9  & 84    & $\mathbf{84.5}$ \\
    \bottomrule
    \end{tabular}%
  \caption{Performance comparison on ImageNet-1K with 224$\times$224 input resolution. F and P denote the FLOPs and number of Params of a model, respectively.
  }
  \label{tab:in1k}%
\end{table}%

\subsection{Network Architecture}
Based on the proposed sparse cross-layer connections, we have developed a new Mamba-based vision backbone named SparX-Mamba, a 4-stage hierarchical architecture with three different variants: tiny, small, and base. In stage 1, we do not include any ganglion layers due to the high computational overhead when large-resolution inputs are used. That is, stage 1 contains only two normal layers. In stage 2, we designate the second layer as a ganglion layer, which has an intra-connection with its preceding layer. Since Stage 3 has more layers, we set different hyperparameters for the three variants. As listed in Table~\ref{tab:model-variants}, due to the relatively small depth of the tiny model, we set \textbf{S} to 2 while \textbf{S} = 3 in both small and base models to avoid high computational cost in deeper models. In stage 4, we make an exception for tiny and small models by setting all layers as ganglion layers, as this stage processes the lowest resolution features and incurs a relatively low computational cost in these models. Regarding more complex base model, only the last layer is set as the ganglion layer to prevent excessive computational costs. Meanwhile, to facilitate communications across different stages, the first ganglion layer of each stage connects to a downsampled version of the final feature from the preceding stage. Furthermore, a cross-layer sliding window is only applied within each stage, to prevent high memory consumption due to excessive feature storage.

\section{Experiments}
\label{sec:experiment}
In this section, we present comprehensive experimental evaluations on representative vision tasks, starting with image classification. Afterwards, the pre-trained models are transferred to downstream tasks, including object detection and semantic segmentation. All experiments were conducted on 8 NVIDIA H800 GPUs. More experimental results are provided in Appendix~\ref{appendix}.

\subsection{Image Classification}
\label{inik_eval}
\textbf{Setup.} We use the ImageNet-1K dataset \cite{deng2009imagenet} and follow the same experimental setting described in DeiT \cite{touvron2021training} for a fair comparison. Our method is compared with many representative vision models: CNN-based models (ConvNeXt \cite{liu2022convnet}, InceptionNeXt \cite{yu2023inceptionnext}, SLaK \cite{liu2022more}, UniRepLKNet \cite{ding2023unireplknet}, and MambaOut \cite{yu2024mambaout}), Transformer-based models (Focal-Transformer \cite{yang2021focal}, PVTv2 \cite{wang2022pvt}, Swin \cite{liu2021swin}, CSWin \cite{dong2022cswin}, and UniFormer \cite{li2022uniformer}), and Mamba-based models (PlainMamba \cite{yang2024plainmamba}, ViM2 \cite{behrouz2024mambamixer}, VMamba \cite{liu2024vmamba}, EfficientVMamba \cite{pei2024efficientvmamba}, LocalVMamba \cite{huang2024localmamba}, and MSVMamba \cite{shi2024multi}).
\par
\textbf{Results.} Table~\ref{tab:in1k} demonstrates the significant advantages of our models over other CNN-, transformer-, and Mamba-based methods. For example, SparX-Mamba-T outperforms ConvNeXt-T/Swin-T by a large margin of 1.4\%/2.2\%, respectively, in top-1 accuracy. Compared to the recent VMamba-T, our model achieves a 1.0\% higher top-1 accuracy with fewer Params and only a slight increase in FLOPs. Furthermore, even with larger models, SparX-Mamba maintains a clear advantage. Meanwhile, we have also provided a more comprehensive comparison with more advanced vision models in Appendix~\ref{appendix}. In general, our method demonstrates competitive performance in ImageNet-1K classification and exhibits clear advantages on downstream tasks.

\begin{table}[!t]
  \centering
  \fontsize{9}{12}\selectfont 
  \setlength{\tabcolsep}{1mm}
    \begin{tabular}{l|cc|cc|cc}
    \toprule
    \multirow{2}[4]{*}{Backbone} & \multirow{2}[4]{*}{F (G)} & \multirow{2}[4]{*}{P (M)} & \multicolumn{2}{c|}{1$\times$ Schedule} & \multicolumn{2}{c}{3$\times$ Schedule} \\
\cmidrule{4-7}          &       &       & $AP_{}^{b}$ & $AP_{}^{m}$ & $AP_{}^{b}$ & $AP_{}^{m}$ \\
    \midrule
    ConvNeXt-T & 262   & 48    & 44.2  & 40.1  & 46.2  & 41.7  \\
    MambaOut-T & 262   & 43    & 45.1  & 41.0  & -  & -  \\
    Focal-T & 291   & 49    & -     & -     & 47.2  & 42.7  \\
    PVTv2-B2 & 309   & 45    & 45.3  & 41.2  & 47.8  & 43.1  \\
    Swin-T & 267   & 48    & 42.7  & 39.3  & 46.0  & 41.6  \\
    CSWin-T & 279   & 42    & 46.7  & 42.2  & 49.0  & 43.6  \\
    UniFormer-S & 269   & 41    & 45.6  & 41.6  & 48.2  & 43.4  \\
    PlainMamba-L2 & 542   & 53    & 46.0  & 40.6  & -     & - \\
    ViM2-T & -     & 39    & 47.1  & 42.4  & -     & - \\
    EfficientVMamba-B & 252   & 53    & 43.7  & 40.2  & 45.0  & 40.8  \\
    LocalVMamba-T & 291   & 45    & 46.7  & 42.2  & 48.7  & 43.4  \\
    MSVMamba-T & 252   & 53    & 46.9  & 42.2  & 48.3  & 43.2  \\
    VMamba-T & 270   & 50    & 47.4  & 42.7  & 48.9  & 43.7  \\
    \rowcolor{gray!20}\textbf{SparX-Mamba-T} & 279   & 47    & $\mathbf{48.1 }$ & $\mathbf{43.1 } $& $\mathbf{50.2 } $& $\mathbf{44.7 }$ \\
    \midrule
    ConvNeXt-S & 348   & 70    & 45.4  & 41.8  & 47.9  & 42.9  \\
    MambaOut-S & 354   & 65    & 47.4  & 42.7  & -  & -  \\
    Focal-S & 401   & 71    & -     & -     & 48.8  & 43.8  \\
    PVTv2-B3 & 397   & 65    & 47.0  & 42.5  & 48.4  & 43.2  \\
    Swin-S & 354   & 69    & 44.8  & 40.9  & 48.2  & 43.2  \\
    CSWin-S & 342   & 54    & 47.9  & 43.2  & 50.0  & 44.5  \\
    UniFormer-B & 399   & 69    & 47.4  & 43.1  & 50.3  & 44.8  \\
    PlainMamba-L3 & 696   & 79    & 46.8  & 41.2  & -     & - \\
    ViM2-S & -     & 62    & 48.5  & 43.1  & -     & - \\
    LocalVMamba-S & 414   & 69    & 48.4  & 43.2  & 49.9  & 44.1  \\
    VMamba-S & 384   & 70    & 48.7  & 43.7  & 49.9  & 44.2  \\
    \rowcolor{gray!20}\textbf{SparX-Mamba-S} & 361   & 67    & $\mathbf{49.4}$ & $\mathbf{44.1}$ & $\mathbf{51.0}$ & $\mathbf{45.2}$ \\
    \midrule
    ConvNeXt-B & 486   & 108   & 47.0  & 42.7  & 48.5  & 43.5  \\
    MambaOut-B & 495   & 100    & 47.4  & 42.0  & -  & -  \\
    Focal-B & 533   & 110   & 45.9  & -     & 49.0  & 43.7  \\
    PVTv2-B5 & 557   & 102   & 47.4  & 42.5  & 48.4  & 42.9  \\
    Swin-B & 496   & 107   & 46.9  & 42.3  & 48.6  & 43.3  \\
    CSWin-B & 526   & 97    & 48.7  & 43.9  & 50.8  & 44.9  \\
    VMamba-B & 485   & 108   & 49.2  & 43.9  & -     & - \\
    \rowcolor{gray!20}\textbf{SparX-Mamba-B} & 498   & 103   & $\mathbf{49.7 }$ & $\mathbf{44.3 }$ & $\mathbf{51.8 }$ & $\mathbf{45.8 }$ \\
    \bottomrule
    \end{tabular}%
    \caption{
    Comparison of object detection and instance segmentation performance on the COCO dataset using Mask R-CNN framework. FLOPs are calculated for the 800$\times$1280 resolution.
    }    
  \label{tab:det}%
\end{table}%

\subsection{Object Detection and Instance Segmentation}
\textbf{Setup.} To evaluate our network architecture on object detection and instance segmentation, we conduct experiments on the COCO 2017 dataset~\cite{lin2014microsoft}. We use the Mask R-CNN framework~\cite{he2017mask} and adopt the same experimental settings as in Swin~\cite{liu2021swin}. The backbone networks are initially pre-trained on ImageNet-1K and subsequently fine-tuned for 12 epochs (1$\times$ schedule) as well as 36 epochs (3$\times$ schedule + multi-scale training). 
\par
\textbf{Results.} As shown in Table~\ref{tab:det}, our models demonstrate superior performance in both object detection and instance segmentation over other models. Specifically, SparX-Mamba-T achieves a 0.7\%/0.4\% higher $AP_{}^{b}$/$AP_{}^{m}$ than VMamba-T when fine-tuned using the 1$\times$ schedule, and surpasses VMamba-T by 1.3\%/1.0\% in $AP_{}^{b}$/$AP_{}^{m}$ when fine-tuned using the 3$\times$ schedule. Notably, SparX-Mamba-T even outperforms VMamba-S, which has significantly more Params and FLOPs, using the 3$\times$ schedule. Meanwhile, our method also demonstrates superior performance compared to other excellent vision backbones.

\begin{table}[!h]
\centering
\setlength{\tabcolsep}{1mm}
\fontsize{9}{12}\selectfont 
\begin{tabular}{l|cccccl}
\toprule
\multirow{2}[4]{*}{Backbone} & \multicolumn{3}{c|}{S-FPN 80K} & \multicolumn{3}{c}{UperNet 160K} \\
\cmidrule{2-7} 
& F (G) & P (M) & \multicolumn{1}{c|}{mIoU} & F (G) & P (M) & mIoU \\
\midrule
ConvNeXt-T & - & - & \multicolumn{1}{c|}{-} & 939 & 60 & 46.0/46.7 \\
InceptionNeXt-T & - & 28 & \multicolumn{1}{c|}{43.1} & 933 & 56 & 47.9/- \\
SLaK-T & - & - & \multicolumn{1}{c|}{-} & 936 & 65 & 47.6/- \\
UniRepLKNet-T & - & - & \multicolumn{1}{c|}{-} & 946 & 61 & 48.6/49.1 \\
MambaOut-T & - & - & \multicolumn{1}{c|}{-} & 938 & 54 & 47.4/48.6 \\
PVTv2-B2 & 167 & 29 & \multicolumn{1}{c|}{45.2} & - & - & - \\
Swin-T & 182 & 32 & \multicolumn{1}{c|}{41.5} & 945 & 60 & 44.5/45.8 \\
CSWin-T & 202 & 26 & \multicolumn{1}{c|}{48.2} & 959 & 59 & 49.3/50.7 \\
UniFormer-S & 247 & 25 & \multicolumn{1}{c|}{46.6} & 1008 & 52 & 47.6/48.5 \\
PlainMamba & - & - & \multicolumn{1}{c|}{-} & 419 & 81 & 49.1/- \\
ViM2-T & - & - & \multicolumn{1}{c|}{-} & - & 51 & 48.6/49.9 \\
EfficientVMamba-B & - & - & \multicolumn{1}{c|}{-} & 930 & 65 & 46.5/47.3 \\
LocalVMamba-T & - & - & \multicolumn{1}{c|}{-} & 970 & 57 & 47.9/49.1 \\
VMamba-T & 189 & 34 & \multicolumn{1}{c|}{47.2$^\dagger$} & 948 & 62 & 48.3/48.6 \\
MSVMamba-T & - & - & \multicolumn{1}{c|}{-} & 942 & 65 & 47.6/48.5 \\
\rowcolor{gray!20}\textbf{SparX-Mamba-T} & 197 & 31 & \multicolumn{1}{c|}{$\textbf{49.5}$} & 954 & 57 & $\textbf{50.0/50.8}$ \\
\midrule
ConvNeXt-S & - & - & \multicolumn{1}{c|}{-} & 1027 & 82 & 48.7/49.6 \\
InceptionNeXt-S & - & 50 & \multicolumn{1}{c|}{45.6} & 1020 & 78 & 50.0/- \\
SLaK-S & - & - & \multicolumn{1}{c|}{-} & 1028 & 91 & 49.4/- \\
UniRepLKNet-S & - & - & \multicolumn{1}{c|}{-} & 1036 & 86 & 50.5/51.0 \\
MambaOut-S & - & - & \multicolumn{1}{c|}{-} & 1032 & 76 & 49.5/50.6 \\
PVTv2-B4 & 291 & 66 & \multicolumn{1}{c|}{47.9} & - & - & - \\
Swin-S & 274 & 53 & \multicolumn{1}{c|}{45.2} & 1038 & 81 & 47.6/49.5 \\
CSWin-S & 271 & 39 & \multicolumn{1}{c|}{49.2} & 1027 & 65 & 50.4/51.5 \\
UniFormer-B & 471 & 54 & \multicolumn{1}{c|}{48.0} & 1227 & 80 & 50.0/50.8 \\
ViM2-S & - & - & \multicolumn{1}{c|}{-} & - & 75 & 50.2/51.4 \\
LocalVMamba-S & - & - & \multicolumn{1}{c|}{-} & 1095 & 81 & 50.0/51.0 \\
VMamba-S & 269 & 54 & \multicolumn{1}{c|}{49.4$^\dagger$} & 1038 & 82 & 50.6/51.2 \\
\rowcolor{gray!20}\textbf{SparX-Mamba-S} & 281 & 51 & \multicolumn{1}{c|}{$\textbf{50.5}$} & 1039 & 77 & $\textbf{51.3}$/$\textbf{52.5}$ \\
\midrule
ConvNeXt-B & - & - & \multicolumn{1}{c|}{-} & 1170 & 122 & 49.1/49.9 \\
InceptionNeXt-B & - & 85 & \multicolumn{1}{c|}{46.4} & 1159 & 115 & 50.6/- \\
SLaK-B & - & - & \multicolumn{1}{c|}{-} & 1172 & 135 & 50.2/- \\
MambaOut-B & - & - & \multicolumn{1}{c|}{-} & 1178 & 112 & 49.6/51.0 \\
PVTv2-B5 & 324 & 91 & \multicolumn{1}{c|}{48.7} & - & - & - \\
Swin-B & 422 & 91 & \multicolumn{1}{c|}{46.0} & 1188 & 121 & 48.1/49.7 \\
CSWin-B & 464 & 81 & \multicolumn{1}{c|}{49.9} & 1222 & 109 & 51.1/52.2 \\
VMamba-B & 409 & 92 & \multicolumn{1}{c|}{49.8$^\dagger$} & 1170 & 122 & 51.0/51.6 \\
\rowcolor{gray!20}\textbf{SparX-Mamba-B} & 422 & 87 & \multicolumn{1}{c|}{$\textbf{51.9}$} & 1181 & 115 & $\textbf{52.3}$/$\textbf{53.4}$ \\
\bottomrule
\end{tabular}%
\caption{Comparison of semantic segmentation performance on the ADE20K dataset. FLOPs are calculated for the 512$\times$2048 resolution. $\dagger$: baselines implemented by ourselves. For UperNet, we report both single-scale and multi-scale mIoU.
}
\label{tab:seg}%
\end{table}

\subsection{Semantic Segmentation}
\label{seg_eval}
\textbf{Setup.} Experiments on semantic segmentation are conducted using the ADE20K dataset~\cite{zhou2017scene}. We employ two segmentation frameworks, Semantic FPN (S-FPN)~\cite{kirillov2019panoptic} and UperNet~\cite{xiao2018unified}. To ensure fair comparisons, we initialize all backbone networks with ImageNet-1K pre-trained weights. Furthermore, we strictly adhere to the same training settings as outlined in previous work \cite{liu2021swin,li2022uniformer}.
\par
\textbf{Results.} As shown in Table~\ref{tab:seg}, our SparX-Mamba demonstrates significant advantages when integrated into both S-FPN and UperNet. For instance, when UperNet is used, SparX-Mamba increases mIoU by 1.7\%/2.2\% over VMamba-T. Performance improvements achieved with small/base models are also evident. Meanwhile, when using S-FPN, SparX-Mamba-T improves VMamba-T by a notable 2.3\% mIoU and outperforms CSWin-T by 1.3\% mIoU. These advantages are also evident in small and base models, effectively demonstrating the robustness of our model in dense prediction tasks.

\subsection{Ablation Studies}
\label{sec:ab_study}
\textbf{Setup.} To assess the effectiveness of individual components in SparX, we conduct comprehensive ablation studies on image classification and semantic segmentation. We first train each model variant on the ImageNet-1K dataset using the same training setting as in Section \ref{inik_eval}. Then, we fine-tune the pre-trained models on the ADE20K dataset, using the S-FPN framework and maintaining an identical training configuration as described in Section \ref{seg_eval}. We have also provided more analytical experiments, which can be found in Appendix \ref{appendix}.

\begin{table}[!t]
  \centering
  \setlength{\tabcolsep}{1mm}
    \fontsize{9}{12}\selectfont 
    \begin{tabular}{c|ccccc|c}
    \toprule
    \multirow{3}[4]{*}{Method} & \multicolumn{5}{c|}{ImageNet-1K}      & ADE20K \\
\cmidrule{2-7}          & Mem     & T     & F     & P     & \multirow{2}[2]{*}{Acc.} & \multirow{2}[2]{*}{mIoU} \\
          &  (MB) & (imgs/s) & (G)   &  (M)  &       &  \\
    \midrule
    VMamba-T & 6784  & 1613  & 4.9   & 31.0  & 82.5  & 47.2 \\
    \midrule
    DGC-Mamba-T &  7523     & 860      & 5.5   & 28.8  & 83.4  & 49.3 \\
    DSN-Mamba-T & 7641  & 830   & 6.0   & 30.9  & 83.4  & \textbf{49.5} \\
    \rowcolor{gray!20}SparX-Mamba-T & 7066  & 1370  & 5.2   & 27.1  & \textbf{83.5} & \textbf{49.5} \\
    \bottomrule
    \end{tabular}%
      \caption{Comparison with DenseNet-like networks. The throughput (T) and memory (Mem) usage are evaluated on a single H800 GPU using a batch size of 128. Due to space constraints, we omit the complexity of the segmentation network. However, it is noteworthy that its complexity is closely aligned with that of the classification network.
      }
  \label{tab:compare_densenet}%
\end{table}%

\par
\textbf{Comparison with DenseNet-like networks.} Based on SparX-Mamba-T, we conduct the following experiments: (1) To validate our sparse cross-layer connections, we maintain the stride (\textbf{S}) = 2 in SparX, and set up cross-layer connections by removing the sliding window and connecting each ganglion layer with all preceding ganglion and normal layers. This design is denoted as ``Dense Ganglion Connections (DGC-Mamba-T)". (2) To verify the superiority of our model over DenseNet~\cite{huang2017densely}, we set \textbf{S}=1 and remove the cross-layer sliding window, thereby designing a network similar to DenseNet. This model is referred to as ``DenseNet-style Network (DSN-Mamba-T)". 
\par
As shown in Table~\ref{tab:compare_densenet}, we observe that both types of dense cross-layer connections (DGC and DSN) incur obvious computational overhead compared to VMamba, particularly in terms of noticeably reduced throughput and increased memory usage, which may hinder scaling to larger models. In contrast, our SparX achieves more pronounced improvements while only slightly reducing throughput and marginally increasing memory usage, demonstrating its high efficiency and effectiveness in Mamba-based models.

\begin{table}[!t]
  \centering
    \setlength{\tabcolsep}{1mm}
    \fontsize{9}{12}\selectfont 
    \begin{tabular}{c|ccccc|c}
    \toprule
    \multirow{3}[4]{*}{Method} & \multicolumn{5}{c|}{ImageNet-1K}      & ADE20K \\
\cmidrule{2-7}          & Mem     & T     & F     & P     & \multirow{2}[2]{*}{Acc.} & \multirow{2}[2]{*}{mIoU} \\
          &  (MB) & (imgs/s) & (G)   &  (M)  &       &  \\
    \midrule
    Concat & 6861  & 1413  & 5.5   & 27.9  & 82.8  & 48.3 \\
    \midrule
    w/o CGCA & 6853  & 1425  & 5.3   & 26.4  & 82.7  & 48.1 \\ 
    CSA   & 7053  & 600   & 5.2   & 26.0  & 83.1  & 49.0 \\
    SRA   & 9333  & 997   & 5.3   & 27.0  & 83.2  & 49.2 \\ 
    SCA   & 12453  & 1275  & 5.2   & 27.9  & 83.3  & 49.2\\
    \rowcolor{gray!20}DMCA  & 7066  & 1370  & 5.2   & 27.1  & \textbf{83.5} & \textbf{49.5} \\
    \bottomrule
    \end{tabular}%
      \caption{Comparison with different cross-layer interaction methods.\vspace{-0mm}}
  \label{tab:compare_dmca}%
\end{table}%

\textbf{Comparison with different interaction methods.} Based on SparX-Mamba-T, we replace our DMCA with other fusion methods for multi-layer feature interactions: 
(1) Remove the entire DMCA module and directly concatenate features from the current layer and selected preceding layers, followed by a Conv-FFN layer~\cite{wang2022pvt} for feature fusion. Using Conv-FFN for feature fusion maintains a comparable level of complexity with the baseline model. This design is referred to as ``Concat". (2) Remove the cross-layer grouped channel attention (CGCA) operation in DMCA. That means there is no need to perform spatial reduction and compute the cross-attention matrix between spatially reduced \textit{query} and \textit{key}, but \textit{value} still needs to be computed. Then \textit{query} and \textit{value} are directly concatenated and fused using a Conv-FFN layer. This design is denoted as ``w/o CGCA". (3) Replace the attention computation in DMCA with Spatial Reduction Attention (SRA) \cite{wang2021pyramid,wang2022pvt} and Cross-Scale Attention (CSA) \cite{shang2023vision} to compare the performance of channel mixing and spatial mixing. (4) Replace our DMCA with spatial cross-attention (SCA) proposed in FcaFormer~\cite{zhang2023fcaformer} to compare the performance of different multi-layer feature aggregation mechanisms. This version is denoted as ``SCA". 
\par
As listed in Table~\ref{tab:compare_dmca}, our DMCA demonstrates better cross-layer feature aggregation and selection thanks to dynamic feature retrieval from preceding layers. Specifically, both ``Concat" and ``w/o CGCA" result in obvious performance degradation compared with our DMCA. Meanwhile, DMCA outperforms both SRA and SCA possibly because DMCA offers dynamic channel mixing, which is complementary to spatial mixing performed with SSMs. It is noteworthy that some previous works have demonstrated that incorporating both channel and spatial dynamics can lead to improved performance \cite{fu2019dual,ding2022davit}. Conversely, both SRA and SCA function as a spatial mixer similar to SSM, compromising the model's ability to effectively represent multidimensional features. More importantly, compared with simple feature concatenation, both SRA and SCA incur higher memory costs and lower throughput due to quadratic complexity, whereas our DMCA only marginally increases computational overhead, indicating greater efficiency.

\begin{table}[!t]
  \centering
    \setlength{\tabcolsep}{1mm}
    \fontsize{9}{12}\selectfont 
    \begin{tabular}{l|ccl|ccl}
    \toprule
    \multicolumn{1}{c|}{\multirow{2}[4]{*}{Method}} & \multicolumn{3}{c|}{ImageNet-1K} & \multicolumn{3}{c}{ADE20K} \\
\cmidrule{2-7}          & F (G) & P (M) & Acc. & F (G) & P (M) & mIoU \\
    \midrule
\rowcolor{gray!20}Baseline & 5.2 & 27.1 & 83.5 & 197 & 30.8 & 49.5 \\
PlainNet & 5.4 & 27.4 & 82.4$_{(-1.1)}$ & 199 & 31.1 & 47.7$_{(-1.8)}$ \\
S=1 & 5.7 & 29.9 & 83.5$_{(+0.0)}$ & 206 & 33.7 & 49.7$_{(+0.2)}$ \\
S=3 & 5.1 & 26.3 & 83.2$_{(-0.3)}$ & 194 & 30.1 & 49.0$_{(-0.5)}$ \\
S=4 & 5.1 & 25.8 & 82.9$_{(-0.6)}$ & 193 & 29.6 & 48.7$_{(-0.8)}$ \\
M=2 & 5.2 & 26.8 & 83.0$_{(-0.5)}$ & 197 & 30.6 & 48.6$_{(-0.9)}$ \\
M=4 & 5.5 & 27.5 & 83.5$_{(+0.0)}$ & 198 & 31.2 & 49.7$_{(+0.2)}$ \\
\bottomrule
\end{tabular}%
\caption{Ablation study on hyperparameters of the SparX.}
\label{tab:ab-connect}%
\end{table}

\begin{table}[t]
    \centering
    \setlength{\tabcolsep}{1mm}
    \fontsize{9}{12}\selectfont 
    \begin{tabular}{l|ccl|ccl}
      \toprule
      \multicolumn{1}{c|}{\multirow{2}[4]{*}{Method}} & \multicolumn{3}{c|}{ImageNet-1K} & \multicolumn{3}{c}{ADE20K} \\
      \cmidrule{2-7}          & F (G) & P (M) & Acc. & F (G) & P (M) & mIoU \\
      \midrule
      \rowcolor{gray!20}Baseline & 5.2   & 27.1  & 83.5  & 197   & 30.8  & 49.5 \\
      w/o DPE & 5.2   & 27.0  & 83.3 $_{(-0.2)}$ & 196   & 30.7  & 49.0 $_{(-0.5)}$ \\
      w/o Skip & 5.2   & 27.0  & 83.1 $_{(-0.4)}$ & 200   & 30.8  & 48.9 $_{(-0.6)}$ \\
      w/o SR & 5.4   & 27.0  & 83.4 $_{(-0.1)}$ & 196   & 30.8  & 49.3 $_{(-0.2)}$ \\
      \bottomrule
    \end{tabular}%
    \caption{Ablation study on DMCA and DPE.\vspace{-0mm}}
    \label{tab:ab-fusion}%
\end{table}

\par
\textbf{Hyperparameter settings in SparX.} We conduct an in-depth exploration of hyperparameter settings in the proposed sparse cross-layer connection mechanism. By using the tiny model as the baseline (i.e., \textbf{S}=2, \textbf{M}=3), we conceive the following model variants: (1) All ganglion layers are replaced with normal layers, i.e., the network no longer has any cross-layer connections. This design is denoted as ``PlainNet". For a fair comparison, we make models considered in this comparison have comparable complexity by adjusting the number of channels in stage 3 and stage 4 of PlainNet to 384 and 576, respectively. (2) Keeping \textbf{M}=3 fixed, we vary the stride hyperparameter (\textbf{S}) controlling the sparsity of ganglion layers to 1, 3, and 4, respectively, to assess the impact of stride on final performance. (3) Keeping \textbf{S}=2 fixed, we investigate the impact of the cross-layer sliding window size by setting \textbf{M} to 2 and 4, respectively.

\textbf{Results.} As shown in Table~\ref{tab:ab-connect}, having dense ganglion layers (i.e., \textbf{S}=1) does not result in an obvious performance improvement even though the model complexity increases. This indicates that our sparse placement of ganglion layers already facilitates cross-layer communication. However, as \textbf{S} further increases, model performance gradually declines, suggesting that larger gaps between ganglion layers adversely affects the performance. Note that although \textbf{S}=3 is the second-best option, considering the increased complexity of deeper models, we set \textbf{S}=3 in our small and base versions for higher computational efficiency. Furthermore, experiments demonstrate that the optimal number of preceding ganglion layers in the cross-layer sliding window is 3, as further increasing \textbf{M} to 4 does not yield any obvious performance gains.

\begin{table*}[!t]
  \centering
    \setlength{\tabcolsep}{1mm}
    \fontsize{9}{12}\selectfont 
\begin{tabular}{c|ccl|ccl|ccllll}
\hline
\multirow{2}{*}{Method} & \multicolumn{3}{c|}{ImageNet-1K} & \multicolumn{3}{c|}{UperNet 160K} & \multicolumn{6}{c}{Mask R-CNN} \\ \cline{2-13} 
& F (G) & P (M) & Acc. & F (G) & P (M) & mIoU  & \multicolumn{1}{c}{F (G)} & \multicolumn{1}{c}{P (M)} & $AP_{1\times}^{b}$ & $AP_{1\times}^{m}$ & $AP_{3\times}^{b}$ & $AP_{3\times}^{m}$ \\ \hline
Swin-T & 4.5   & 29    & 81.3 & 945   & 60    & 44.5 & 48 & 267 & 42.7  & 39.3  & 46.0  & 41.6  \\
\rowcolor{gray!20}\textbf{SparX-Swin-T} & 4.7   & 26    & \textbf{82.6}$_{(+1.3)}$ & 948   & 57 &\textbf{45.5}$_{(+1.0)}$ & 45                        & 270                       &       \textbf{44.1}$_{(+1.4)}$&       \textbf{40.8}$_{(+1.5)}$&       \textbf{47.6}$_{(+1.6)}$&       \textbf{43.0}$_{(+1.4)}$\\ \hline
Swin-B & 15.4  & 88    & 83.5                                      & 1188  & 121   & 48.1                                      & 107                       & 496                       & 46.9  & 42.3  & 48.6  & 43.3  \\
\rowcolor{gray!20}\textbf{SparX-Swin-B} & 16.2  & 88    & \textbf{83.9}$_{(+0.4)}$ & 1201  & 118   & \textbf{48.6}$_{(+0.5)}$ & 107                       & 518                       &       \textbf{48.1}$_{(+1.2)}$&       \textbf{42.9}$_{(+0.6)}$&       \textbf{49.1}$_{(+0.5)}$ &       \textbf{43.7}$_{(+0.4)}$ \\ \hline
\end{tabular}
  \caption{Performance Comparison when SparX is applied to Vision Transformers.
  }
  \label{tab:ab_mixers}%
\end{table*}

\par
\textbf{Ablations of DMCA and DPE.} By using the tiny model as the baseline, we further investigate the effect of DMCA and DPE~\cite{chu2022conditional} by evaluating the performance of the following alternative model designs. (1) Remove the DPE module, and the resulting model is denoted as ``w/o DPE". (2) Investigate the necessity of concatenating the output of GCCA ($\mathbf{Z}$) with \textit{query} and \textit{value} by only using the output of GCCA without concatenating it with \textbf{X} and $\mathbf{Y_v}$. This design is denoted as ``w/o Skip". (3) Eliminate the spatial reduction (SR) operations, and use the original spatial resolution of \textit{query} and \textit{key} when computing GCCA. This design is denoted as ``w/o SR". According to the results given in Table~\ref{tab:ab-fusion}, the use of DPE boosted the top-1 accuracy by 0.2\% and mIoU by 0.5\%. Then, we can find that ``w/o Skip" also gives rise to a performance drop, reflecting the importance of the shortcut connection. Moreover, although spatial reduction (SR) compresses spatial tokens, it actually slightly improves performance. One possible explanation is that utilizing compressed information is sufficient when estimating image-level feature similarities.

\begin{table}[!b]
  \centering
  \setlength{\tabcolsep}{1mm}
    \fontsize{9}{12}\selectfont 
    \begin{tabular}{l|ccl|ccl}
    \toprule
    \multirow{2}[4]{*}{Method} & \multicolumn{3}{c|}{ImageNet-1K} & \multicolumn{3}{c}{UperNet 160K} \\
\cmidrule{2-7}          & F (G) & P (M) & Acc. & F (G) & P (M) & mIoU \\
    \midrule
    VMamba-T &       &       &       &       &       &  \\
    (SS2D) & 4.9   & 31    & 82.5  & 948  & 62    & 48.3 \\
    \midrule
    \rowcolor{gray!6}\textbf{SparX-Mamba} &       &       &       &       &       &  \\
    \rowcolor{gray!6}\textbf{-T (SSM)} & 4.8   & 25    & 82.9$_{(+0.4)}$  & 946   & 55    & 48.1$_{(-0.2)}$ \\
    \rowcolor{gray!13}\textbf{SparX-Mamba} &       &       &       &       &       &  \\
    \rowcolor{gray!13}\textbf{-T (Bi-SSM)} & 5.1   & 25    & 83.1$_{(+0.6)}$  & 951   & 55    & 48.7$_{(+0.4)}$ \\
    \rowcolor{gray!20}\textbf{SparX-Mamba} &       &       &       &       &       &  \\
    \rowcolor{gray!20}\textbf{-T (SS2D)} & 5.2   & 27    & 83.5$_{(+1.0)}$  & 954   & 57    & 50.0$_{(+1.7)}$ \\
    \bottomrule
    \end{tabular}%
    \caption{Comparison of image classification and semantic segmentation performance when SparX is applied to various SSM modules.}
\label{tab:ab_ssm}%
\end{table}%

\subsection{Versatility Analysis}
\textbf{Setup.} To demonstrate the versatility of our cross-layer connectivity pattern, we have conducted experiments by incorporating SparX into networks with different token mixer architectures, including self-attention, vanilla SSM in Mamba, and Bi-SSM in ViM. To ensure a fair comparison with other hierarchical networks, we follow the conventional design, where a token mixer is followed by an FFN, with the token mixer having the aforementioned multiple options. Specifically, taking SparX-Mamba-T as a basic network, we replace the SS2D block with shifted window attention proposed in Swin Transformer \cite{liu2021swin} to obtain a network denoted as SparX-Swin-T (i.e., {\pcr{Shifted Window Attention}} $\rightarrow$ {\pcr{FFN}}), and increase the number of channels in stage 4 from 512 to 576 to align its computational cost with Swin-T. Next, we scale up the model size of SparX-Swin-T by increasing the depths to [2, 2, 27, 3] and channels to [120, 240, 396, 616] to obtain a larger model, denoted as SparX-Swin-B, which has a model complexity comparable to that of Swin-B. Moreover, we also replace the SS2D module in VSS with the vanilla SSM module in original Mamba \cite{gu2023mamba} (i.e., {\pcr{SSM}} $\rightarrow$ {\pcr{FFN}}) and Bi-SSM module in ViM \cite{zhu2024vision} (i.e., {\pcr{Bi-SSM}} $\rightarrow$ {\pcr{FFN}}). For fair comparisons, we have strictly maintained the similarity of other micro-blocks, such as removing DPE and using identical patch embedding layers. 
\par
\textbf{Results.} As shown in Table~\ref{tab:ab_mixers}, SparX-Swin-T/B models outperform their vanilla counterparts on both image classification and dense prediction tasks while maintaining similar computational costs. This demonstrates that our method is general and equally applicable to Transformer-based models. Meanwhile, as shown in Table~\ref{tab:ab_ssm}, although vanilla SSM and Bi-SSM do not utilize the DWConv employed in SS2D of VMamba, our SparX enables both vanilla SSM and Bi-SSM to outperform VMamba-T in terms of accuracy and computational efficiency. On the semantic segmentation task, the performance of SparX-Mamba-T (SSM) marginally lags behind VMamba-T. This is because causal modeling in vanilla SSM fails to capture sufficient contextual information. When the Bi-SSM module in ViM is used instead, the mIoU improves significantly by 0.4\%, demonstrating better performance than VMamba-T.

\section{Conclusion}
In this work, we propose a new skip-connection strategy named SparX, drawing inspiration from the Retinal Ganglion Cell (RGC) layer in the human visual system. SparX aims to create sparse cross-layer connections to enhance information flow and promote feature distillation and reuse in vision backbone networks. In addition, we propose a Dynamic Multi-layer Channel Aggregator (DMCA) that facilitates dynamic feature aggregation and interaction across layers. Based on SparX, we further propose SparX-Mamba and SparX-Swin, both of which demonstrate superior performance across a range of challenging vision tasks.

\appendix
\section{Appendix}
\label{appendix}

\begin{table*}[thb]
  \centering
  \setlength{\tabcolsep}{1mm}
    \fontsize{9}{12}\selectfont 
    \resizebox{0.75\textwidth}{!}{
    \begin{tabular}{l|cc|cccccc|cccccc}
    \toprule
    \multirow{2}[4]{*}{Backbone} & \multirow{2}[4]{*}{F (G)} & \multirow{2}[4]{*}{P (M)} & \multicolumn{6}{c|}{Mask R-CNN 1$\times$ Schedule}   & \multicolumn{6}{c}{Mask R-CNN 3$\times$+MS Schedule} \\
\cmidrule{4-15}          &       &       & $AP_{}^{b}$ & $AP_{50}^{b}$ & $AP_{75}^{b}$ & $AP_{}^{m}$ & $AP_{50}^{m}$ & $AP_{75}^{m}$ & $AP_{}^{b}$ & $AP_{50}^{b}$ & $AP_{75}^{b}$ & $AP_{}^{m}$ & $AP_{50}^{m}$ & $AP_{75}^{m}$ \\
    \midrule
    ConvNeXt-T & 262   & 48    & 44.2  & 66.6  & 48.3  & 40.1  & 63.3  & 42.8  & 46.2  & 67.9  & 50.8  & 41.7  & 65.0  & 44.9  \\
    Focal-T & 291   & 49    & -     & -     & -     & -     & -     & -     & 47.2  & 69.4  & 51.9  & 42.7  & 66.5  & 45.9  \\
    PVTv2-B2 & 309   & 45    & 45.3  & 67.1  & 49.6  & 41.2  & 64.2  & 44.4  & 47.8  & 69.7  & 52.6  & 43.1  & 66.8  & 46.7  \\
    Swin-T & 267   & 48    & 42.7  & 65.2  & 46.8  & 39.3  & 62.2  & 42.2  & 46.0  & 68.1  & 50.3  & 41.6  & 65.1  & 44.9  \\
    CSWin-T & 279   & 42    & 46.7  & 68.6  & 51.3  & 42.2  & 65.6  & 45.4  & 49.0  & 70.7  & 53.7  & 43.6  & 67.9  & 46.6  \\
    UniFormer-S & 269   & 41    & 45.6  & 68.1  & 49.7  & 41.6  & 64.8  & 45.0  & 48.2  & 70.4  & 52.5  & 43.4  & 67.1  & 47.0  \\
    PlainMamba-L2 & 542   & 53    & 46.0  & 66.9  & 50.1  & 40.6  & 63.8  & 43.6  & -     & -     & -     & -     & -     & - \\
    ViM2-T & -     & 39    & 47.1  & 68.7  & 50.9  & 42.4  & 65.6  & 45.5  & -     & -     & -     & -     & -     & - \\
    EfficientVMamba-B & 252   & 53    & 43.7  & 66.2  & 47.9  & 40.2  & 63.3  & 42.9  & 45.0  & 66.9  & 49.2  & 40.8  & 64.1  & 43.7  \\
    LocalVMamba-T & 291   & 45    & 46.7  & 68.7  & 50.8  & 42.2  & 65.7  & 45.5  & 48.7  & 70.1  & 53.0  & 43.4  & 67.0  & 46.4  \\
    VMamba-T & 270   & 50    & 47.4  & 69.5  & 52.0  & 42.7  & 66.3  & 46.0  & 48.9  & 70.6  & 53.6  & 43.7  & 67.7  & 46.8  \\
    \rowcolor{gray!20}\textbf{SparX-Mamba-T} & 279   & 47    & $\mathbf{48.1}$ & $\mathbf{70.1}$ & $\mathbf{52.8}$ & 
    $\mathbf{43.1}$ & $\mathbf{67.0}$ & $\mathbf{46.5}$ & $\mathbf{50.2}$ & $\mathbf{71.8}$ & $\mathbf{55.2}$ & $\mathbf{44.7}$ & $\mathbf{68.8}$ & $\mathbf{48.4}$ \\
    \midrule
    ConvNeXt-S & 348   & 70    & 45.4  & 67.9  & 50.0  & 41.8  & 65.2  & 45.1  & 47.9  & 70.0  & 52.7  & 42.9  & 66.9  & 46.2  \\
    Focal-S & 401   & 71    & -     & -     & -     & -     & -     & -     & 48.8  & 70.5  & 53.6  & 43.8  & 67.7  & 47.2  \\
    PVTv2-B3 & 397   & 65    & 47.0  & 68.1  & 51.7  & 42.5  & 65.7  & 45.7  & 48.4  & 69.8  & 53.3  & 43.2  & 66.9  & 46.7  \\
    Swin-S & 354   & 69    & 44.8  & 66.6  & 48.9  & 40.9  & 63.4  & 44.2  & 48.2  & 69.8  & 52.8  & 43.2  & 67.0  & 46.1  \\
    CSWin-S & 342   & 54    & 47.9  & 70.1  & 52.6  & 43.2  & 67.1  & 46.2  & 50.0  & 71.3  & 54.7  & 44.5  & 68.4  & 47.7  \\
    UniFormer-B & 399   & 69    & 47.4  & 69.7  & 52.1  & 43.1  & 66.0  & 46.5  & 50.3  & 72.7  & 55.3  & 44.8  & 69.0  & 48.3  \\
    PlainMamba-L3 & 696   & 79    & 46.8  & 68.0  & 51.1  & 41.2  & 64.7  & 43.9  & -     & -     & -     & -     & -     & - \\
    ViM2-S & -     & 62    & 48.5  & 69.9  & 52.8  & 43.1  & 66.8  & 46.5  & -     & -     & -     & -     & -     & - \\
    LocalVMamba-S & 414   & 69    & 48.4  & 69.9  & 52.7  & 43.2  & 66.7  & 46.5  & 49.9  & 70.5  & 54.4  & 44.1  & 67.8  & 47.4  \\
    VMamba-S & 384   & 70    & 48.7  & 70.0  & 53.4  & 43.7  & 67.3  & 47.0  & 49.9  & 70.9  & 54.7  & 44.2  & 68.2  & 47.7  \\
    \rowcolor{gray!20}\textbf{SparX-Mamba-S} & 361   & 67    & $\mathbf{49.4}$ & $\mathbf{71.1}$ & $\mathbf{54.2}$ & $\mathbf{44.1}$ & $\mathbf{68.3}$ & $\mathbf{47.7}$ & $\mathbf{51.0}$ & $\mathbf{71.9}$ & $\mathbf{55.7}$ & $\mathbf{45.2}$ & $\mathbf{69.3}$ & $\mathbf{48.8}$ \\
    \midrule
    ConvNeXt-B & 486   & 108   & 47.0  & 69.4  & 51.7  & 42.7  & 66.3  & 46.0  & 48.5  & 70.1  & 53.3  & 43.5  & 67.1  & 46.7  \\
    Focal-B & 533   & 110   & 45.9  & -     & -     & -     & -     & -     & 49.0  & 70.1  & 53.6  & 43.7  & 67.6  & 47.0  \\
    PVTv2-B5 & 557   & 102   & 47.4  & 68.6  & 51.9  & 42.5  & 65.7  & 46.0  & 48.4  & 69.2  & 52.9  & 42.9  & 66.6  & 46.2  \\
    Swin-B & 496   & 107   & 46.9  & -     & -     & 42.3  & -     & -     & 48.6  & 70.0  & 53.4  & 43.3  & 67.1  & 46.7  \\
    CSWin-B & 526   & 97    & 48.7  & 70.4  & 53.9  & 43.9  & 67.8  & 47.3  & 50.8  & 72.1  & 55.8  & 44.9  & 69.1  & 48.3  \\
    VMamba-B & 485   & 108   & 49.2  & 70.9  & 53.9  & 43.9  & 67.7  & 47.6  & -     & -     & -     & -     & -     & - \\
    \rowcolor{gray!20}\textbf{SparX-Mamba-B} & 498   & 103   & $\mathbf{49.7}$ & $\mathbf{71.6}$ & $\mathbf{54.6}$ & $\mathbf{44.3}$ & $\mathbf{68.4}$ & $\mathbf{48.0}$ & $\mathbf{51.8}$ & $\mathbf{73.1}$ & $\mathbf{56.4}$ & $\mathbf{45.8}$ & $\mathbf{70.2}$ & $\mathbf{49.6}$ \\
    \bottomrule
    \end{tabular}%
   }
    \caption{Performance comparison of Mask R-CNN (the complete version of Table~\ref{tab:det}).}
  \label{tab:det_appendix}%
\end{table*}%

\subsection{More Detailed Comparisons in Object Detection and Instance Segmentation}
\label{sec:det_append}
We provide a more comprehensive performance comparison on object detection and instance segmentation by introducing metrics based on different IoU thresholds, which serves as a supplement to Table~\ref{tab:det}. The results are presented in Table~\ref{tab:det_appendix}, and it is evident that our model outperforms other methods across all metrics.

\begin{table}[h]
\centering
\setlength{\tabcolsep}{1mm}
\fontsize{9}{12}\selectfont 
\resizebox{0.25\textwidth}{!}{
\begin{tabular}{lccc}
\toprule
Method & F (G) & P (M) & Acc. (\%) \\
\midrule
Conv2Former-T & 4.4 & 27 & 83.2 \\
InternImage-T & 5.0 & 30 & 83.5 \\
PeLK-T & 5.6 & 29 & 82.6 \\
CMT-S & 4.0 & 25 & 83.5 \\
MaxViT-T & 5.6 & 31 & 83.7 \\
BiFormer-S & 4.5 & 26 & 83.8 \\
MPViT-S & 4.7 & 23 & 83.0 \\
NAT-T & 4.3 & 28 & 83.2 \\
CrossFormer++-S & 4.4 & 23 & 83.2 \\
ViM-S & - & 26 & 81.6 \\
ViM2-T & - & 20 & 82.7 \\
PlainMamba-L2 & 8.1 & 25 & 81.6 \\
EfficientVMamba-B & 4.0 & 33 & 81.8 \\
VMamba-T & 4.9 & 31 & 82.5 \\
LocalVMamba-T & 5.7 & 26 & 82.7 \\
\rowcolor{gray!20}\textbf{SparX-Mamba-T} & 5.2 & 27 & 83.5 \\
\midrule
Conv2Former-S & 8.7 & 50 & 84.1 \\
InternImage-S & 8.0 & 50 & 84.2 \\
PeLK-S & 10.7 & 50 & 83.9 \\
MaxViT-S & 11.7 & 69 & 84.5 \\
BiFormer-B & 9.8 & 57 & 84.3 \\
NAT-S & 7.8 & 51 & 83.7 \\
CrossFormer++-B & 9.5 & 52 & 84.2 \\
ViM2-S & - & 43 & 83.7 \\
VMamba-S & 8.7 & 50 & 83.6 \\
LocalVMamba-S & 11.4 & 50 & 83.7 \\
\rowcolor{gray!20}\textbf{SparX-Mamba-S} & 9.3 & 47 & 84.2 \\
\midrule
Conv2Former-B & 15.9 & 90 & 84.4 \\
InternImage-B & 16.0 & 97 & 84.9 \\
PeLK-B & 18.3 & 89 & 84.2 \\
MPViT-B & 16.4 & 75 & 84.3 \\
NAT-B & 13.7 & 90 & 84.3 \\
CrossFormer++-L & 16.6 & 92 & 84.7 \\
ViM2-B & - & 74 & 83.9 \\
VMamba-B & 15.4 & 89 & 83.9 \\
\rowcolor{gray!20}\textbf{SparX-Mamba-B} & 15.9 & 84 & 84.5 \\
\bottomrule
\end{tabular}
}
\caption{Comparison with more advanced vision backbones on the ImageNet-1K dataset.}
\label{tab:in1k-p2}
\vspace{-5pt}
\end{table}

\begin{table}[!h]
  \centering
    \setlength{\tabcolsep}{1mm}
    \fontsize{9}{12}\selectfont 
    \resizebox{0.4\textwidth}{!}{
    \begin{tabular}{l|ccc|cccc}
    \toprule
    \multirow{2}[4]{*}{Backbone} & \multicolumn{3}{c|}{S-FPN 80K} & \multicolumn{4}{c}{UperNet 160K} \\
\cmidrule{2-8}          & F (G) & P (M) & mIoU  & F (G) & P (M) & mIoU  & MS-mIoU \\
    \midrule
    Conv2Former-T & -     & -     & -     & 946   & 59    & 45.8  & -  \\
    InternImage-T & -     & -     & -     & 944   & 59    & 47.9  & 48.1  \\
    PeLK-T & -     & -     & -     & 970  & 62    & 48.1  & -  \\
    BiFormer-S & -     & -     & 48.9  & -     & -     & 49.8  & $\mathbf{50.8}$  \\
    MPViT-S & -     & -     & -     & 943   & 52    & 48.3  & - \\
    NAT-T & -     & -     & -     & 934   & 58    & 47.1  & 48.4  \\
    CrossFormer++-S & 200   & 27    & 47.4  & 964   & 53    & 49.4  & 50.0  \\
    PlainMamba-L3 & -     & -     & -     & 419   & 81    & 49.1  & - \\
    ViM2-T & -     & -     & -     & -     & 51    & 48.6  & 49.9  \\
    EfficientVMamba-B & -     & -     & -     & 930   & 65    & 46.5  & 47.3  \\
    LocalVMamba-T & -     & -     & -     & 970   & 57    & 47.9  & 49.1  \\
    VMamba-T & 189   & 34    & 47.2  & 948   & 62    & 48.3  & 48.6  \\
    \rowcolor{gray!20}\textbf{SparX-Mamba-T} & 197   & 31    & $\mathbf{49.5}$  & 954   & 57    & $\mathbf{50.0}$  &$\mathbf{50.8}$  \\
    \midrule
    Conv2Former-S & -     & -     & -     & 1021   & 78    &50.3  & -  \\
    InternImage-S & -     & -     & -     & 1017  & 80    & 50.1  & 50.9  \\
    PeLK-S & -     & -     & -     & 1077  & 84    & 49.7  & -  \\
    BiFormer-B & -     & -     & 49.9  & -     & -     & {51.0}  & 51.7  \\
    NAT-S & -     & -     & -     & 1071  & 82    & 48.0  & 49.5  \\
    CrossFormer++-B & 331   & 56    & 48.6  & 1090  & 84    & 50.7  & 51.0  \\
    ViM2-S & -     & -     & -     & -     & 75    & 50.2  & 51.4  \\
    LocalVMamba-S & -     & -     & -     & 1095  & 81    & 50.0  & 51.0  \\
    VMamba-S & 269   & 54    & 49.4  & 1039  & 82    & 50.6  & 51.2  \\
    \rowcolor{gray!20}\textbf{SparX-Mamba-S} & 281   & 51    & $\mathbf{50.5}$ & 1039  & 77    & $\mathbf{51.3}$ &$\mathbf{52.5}$  \\
    \midrule
    Conv2Former-B & -     & -     & -     & 1171   & 119    &51.0  & -  \\
    InternImage-B & -     & -     & -     & 1185  & 128   & 50.8  & 51.3  \\
    PeLK-B & -     & -     & -     & 1237  & 126 & 50.4  & -  \\
    MPViT-B & -     & -    & -  & 1186  & 105   & 50.3  & - \\
    CrossFormer++-L & 483   & 96    & 49.5  & 1258  & 126   & 51.0  & 51.9  \\
    VMamba-B & 409   & 92    & 49.8  & 1170  & 122   & 51.0  & 51.6  \\
    \rowcolor{gray!20}\textbf{SparX-Mamba-B} & 422   & 87    & $\mathbf{51.9}$  & 1181  & 115   & $\mathbf{52.3} $ &$\mathbf{53.4}$  \\
    \bottomrule
    \end{tabular}%
}
  \caption{Comparison with more advanced vision backbones on the ADE20K dataset for semantic segmentation.}
  \label{tab:seg-p2}%
\end{table}

\par
\begin{table*}[t]
  \centering
  \setlength{\tabcolsep}{1mm}
    \fontsize{9}{12}\selectfont 
  \resizebox{0.75\textwidth}{!}{
    \begin{tabular}{l|cc|cccccc|cccccc}
    \toprule
    \multirow{2}[4]{*}{Backbone} & \multirow{2}[4]{*}{F (G)} & \multirow{2}[4]{*}{P (M)} & \multicolumn{6}{c|}{Mask R-CNN 1$\times$ Schedule}   & \multicolumn{6}{c}{Mask R-CNN 3$\times$ + MS Schedule} \\
\cmidrule{4-15}          &       &       & $AP_{}^{b}$ & $AP_{50}^{b}$ & $AP_{75}^{b}$ & $AP_{}^{m}$ & $AP_{50}^{m}$ & $AP_{75}^{m}$ & $AP_{}^{b}$ & $AP_{50}^{b}$ & $AP_{75}^{b}$ & $AP_{}^{m}$ & $AP_{50}^{m}$ & $AP_{75}^{m}$ \\
    \midrule
    Conv2Former-T & 255     &48  & -     & -     & -     & -     & -     & -   & 48.0  & 69.5  & 52.7  & 43.0  &66.8  &46.1  \\
    InternImage-T & 270   & 49    & 47.2  & 69.0  & 52.1  & 42.5  & 66.1  & 45.8  & 49.1  & 70.4  & 54.1  & 43.7  & 67.3  & 47.3  \\
    CMT-S & 249   & 45    & 44.6  & 66.8  & 48.9  & 40.7  & 63.9  & 43.4  & -     & -     & -     & -     & -     & - \\
    BiFormer-S & -     & -     & 47.8  & 69.8  & 52.3  & $\mathbf{43.2}$  & 66.8  & 46.5  & -     & -     & -     & -     & -     & - \\
    MPViT-S & 268   & 43    & 46.4     & 68.6     & 51.2     & 42.4     &65.6     & 45.7     & 48.4  & 70.5  & 52.6  & 43.9  & 67.6  & 47.5  \\
    NAT-T & 258   & 48    & -     & -     & -     & -     & -     & -     & 47.8  & 69.0  & 52.6  & 42.6  & 66.0  & 45.9  \\
    CrossFormer++-S & 287   & {43} & 46.4  & 68.8  & 51.3  & 42.1  & 65.7  & 45.4  & 49.5  & 71.6  & 54.1  & 44.3  & 68.5  & 47.6 \\
    PlainMamba-L2 & 542   & 53    & 46.0  & 66.9  & 50.1  & 40.6  & 63.8  & 43.6  & -     & -     & -     & -     & -     & - \\
    ViM2-T & -     & 39    & 47.1  & 68.7  & 50.9  & 42.4  & 65.6  & 45.5  & -     & -     & -     & -     & -     & - \\
    EfficientVMamba-B & 252   & 53    & 43.7  & 66.2  & 47.9  & 40.2  & 63.3  & 42.9  & 45.0  & 66.9  & 49.2  & 40.8  & 64.1  & 43.7  \\
    LocalVMamba-T & 291   & 45    & 46.7  & 68.7  & 50.8  & 42.2  & 65.7  & 45.5  & 48.7  & 70.1  & 53.0  & 43.4  & 67.0  & 46.4  \\
    VMamba-T & 270   & 50    & 47.4  & 69.5  & 52.0  & 42.7  & 66.3  & 46.0  & 48.9  & 70.6  & 53.6  & 43.7  & 67.7  & 46.8  \\
    \rowcolor{gray!20}\textbf{SparX-Mamba-T} & 279   & 47    & $\mathbf{48.1}$ & $\mathbf{70.1}$ & $\mathbf{52.8}$ & $\mathrm{43.1}$ & $\mathbf{67.0} $& $\mathbf{46.5}$ & $\mathbf{50.2}$ & $\mathbf{71.8}$ & $\mathbf{55.2} $& $\mathbf{44.7}$ & $\mathbf{68.8}$ & $\mathbf{48.4} $\\
    \midrule
    InternImage-S & 340   & 69    & 47.8  & 69.8  & 52.8  & 43.3  & 67.1  & 46.7  & 49.7  & 71.1  & 54.5  & 44.5  & 68.5  & 47.8  \\
    BiFormer-B & -     & -     & 48.6  & 70.5  & 53.8  & 43.7  & 67.6  & 47.1  & -     & -     & -     & -     & -     & - \\
    NAT-S & 330   & 70    & -     & -     & -     & -     & -     & -     & 48.4  & 69.8  & 53.2  & 43.2  & 66.9  & 46.4  \\
    CrossFormer++-B & 408   & 72  & 47.7  & 70.2  & 52.7  & 43.2  & 67.3  & 46.7  & 50.2  & 71.8  & 54.9  & 44.6  & 68.7  & 48.1 \\
    PlainMamba-L3 & 696   & 79    & 46.8  & 68.0  & 51.1  & 41.2  & 64.7  & 43.9  & -     & -     & -     & -     & -     & - \\
    ViM2-S & -     & 62    & 48.5  & 69.9  & 52.8  & 43.1  & 66.8  & 46.5  & -     & -     & -     & -     & -     & - \\
    LocalVMamba-S & 414   & 69    & 48.4  & 69.9  & 52.7  & 43.2  & 66.7  & 46.5  & 49.9  & 70.5  & 54.4  & 44.1  & 67.8  & 47.4  \\
    VMamba-S & 384   & 70    & 48.7  & 70.0  & 53.4  & 43.7  & 67.3  & 47.0  & 49.9  & 70.9  & 54.7  & 44.2  & 68.2  & 47.7  \\
    \rowcolor{gray!20}\textbf{SparX-Mamba-S} & 361   & 67    & $\mathbf{49.4}$ & $\mathbf{71.1}$ & $\mathbf{54.2}$ & $\mathbf{44.1}$ & $\mathbf{68.3}$ & $\mathbf{47.7}$ & $\mathbf{51.0}$      & $\mathbf{71.9}$      &$\mathbf{55.7}$       &$\mathbf{45.2}$       &$\mathbf{69.3}$       &$\mathbf{48.8}$  \\
    \midrule
    InternImage-B & 501   & 115   & 48.8  & 70.9  & {54.0}  & {44.0} & 67.8  & 47.4  & 50.3  & 71.4  & 55.3  & 44.8  & 68.7  & 48.0  \\
    MPViT-B & 503   & 95    & 48.2     & 70.0     & 52.9     & 43.5     & 67.1    & 46.8     & 49.5  & 70.9  & 54.0  & 44.5  & 68.3  & 48.3  \\
    VMamba-B & 485   & 108   & 49.2  & 70.9  & 53.9  & 43.9  & 67.7  & {47.6}  & -     & -     & -     & -     & -     & - \\
    \rowcolor{gray!20}\textbf{SparX-Mamba-B} & 498 & 103 & $\mathbf{49.7}$ & $\mathbf{71.6}$ & $\mathbf{54.6}$ & $\mathbf{44.3}$ & $\mathbf{68.4}$ & $\mathbf{48.0}$ & $\mathbf{51.8} $ &$\mathbf{73.1}$ & $\mathbf{56.4} $ & $\mathbf{45.8} $& $\mathbf{70.2}$ & $\mathbf{49.6}$ \\    
    \bottomrule
    \end{tabular}%
}
  \caption{Comparison with more advanced vision backbones on the COCO 2017 dataset for object detection.}
  \label{tab:det-p2}%
\end{table*}%

\begin{table}[!h]
  \centering
  \setlength{\tabcolsep}{1mm}
\fontsize{9}{12}\selectfont 
  \resizebox{0.325\textwidth}{!}{
    \begin{tabular}{lcccc}
    \toprule
    Image Size & ConvNeXt-T & Swin-T & VMamba-T & \textbf{SparX-Mamba-T} \\
    \midrule
    \textbf{224$\times$224} &       &       &       &  \\
    FLOPs & 4.5   & 4.5   & 4.9   & 5.2  \\
    Params & 29.0  & 28.0  & 31.0  & 27.1  \\
    \rowcolor{gray!20}Acc. (\%) & 82.1  & 81.3  & 82.5  & $\mathbf{83.5}$ \\
    \midrule
    \textbf{384$\times$384} &       &       &       &  \\
    FLOPs & 13.1  & 14.5  & 14.3  & 15.4  \\
    Params & 29.0  & 28.0  & 31.0  & 27.1  \\
    \rowcolor{gray!20}Acc. (\%) & 81.0  & 80.7  & 82.5  & $\mathbf{84.0} $\\
    \midrule
    \textbf{512$\times$512} &       &       &       &  \\
    FLOPs & 23.3  & 26.6  & 25.4  & 27.4  \\
    Params & 29.0  & 28.0  & 31.0  & 27.1  \\
    \rowcolor{gray!20}Acc. (\%) & 78.0  & 79.0  & 81.1  & $\mathbf{82.9} $\\
    \midrule
    \textbf{640$\times$640} &       &       &       &  \\
    FLOPs & 36.5  & 45.0  & 39.6  & 42.8  \\
    Params & 29.0  & 28.0  & 31.0  & 27.1  \\
    \rowcolor{gray!20}Acc. (\%) & 74.3  & 76.6  & 79.3  & $\mathbf{81.5}$ \\
    \midrule
    \textbf{768$\times$768} &       &       &       &  \\
    FLOPs & 52.5  & 70.7  & 57.1  & 61.7  \\
    Params & 29.0  & 28.0  & 31.0  & 27.1  \\
    \rowcolor{gray!20}Acc. (\%) & 69.5  & 73.1  & 76.1  & $\mathbf{79.5}$ \\
    \midrule
    \textbf{1024$\times$1024} &       &       &       &  \\
    FLOPs & 93.3  & 152.5  & 101.5  & 109.6  \\
    Params & 29.0  & 28.0  & 31.0  & 27.1  \\
    \rowcolor{gray!20}Acc. (\%) & 55.4  & 61.9  & 62.3  & $\mathbf{71.8}$ \\
    \bottomrule
    \end{tabular}%
}
\caption{Comparison of generalization ability over an increasing input resolution}
  \label{tab:resolutions}%
\end{table}%

\begin{table}[!h]
  \centering
    \setlength{\tabcolsep}{1mm}
\fontsize{9}{12}\selectfont 
    \resizebox{0.325\textwidth}{!}{
    \begin{tabular}{lcccc}
    \toprule
    Method & F (G) & P (M) & T (imgs/s) & Acc. (\%) \\
    \midrule
    ConvNeXt-T & 4.5   & 29    & 2359  & 82.1  \\
    Swin-T & 4.5   & 29    & 2416  & 81.3  \\
    Focal-T & 4.9   & 29    & 953   & 82.2  \\
    BiFormer-S & 4.5   & 26    & 1072  & 83.8  \\
    MPViT-S & 4.7   & 23    & 1534  & 83.0  \\
    EfficientVMamba-B & 4.0   & 33    & 1943  & 81.8  \\
    VMamba-T & 4.9   & 31    & 1613  & 82.5  \\
    LocalVMamba-T & 5.7   & 26    & 597   & 82.7  \\
    \rowcolor{gray!20}\textbf{SparX-Mamba-T} & 5.2   & 27    & 1370  & 83.5  \\
    \midrule
    ConvNeXt-S & 8.7   & 50    & 1359  & 83.1  \\
    Swin-S & 8.7   & 50    & 1419  & 83.0  \\
    Focal-S & 9.4   & 51    & 558   & 83.6  \\
    BiFormer-B & 9.8   & 57    & 673   & 84.3  \\
    VMamba-S & 8.7   & 50    & 1021  & 83.6  \\
    LocalVMamba-S & 11.4  & 50    & 337   & 83.7  \\
    \rowcolor{gray!20}\textbf{SparX-Mamba-S} & 9.3   & 47    & 871   & 84.2  \\
    \midrule
    ConvNeXt-B & 15.4  & 89    & 957   & 83.8  \\
    Swin-B & 15.4  & 88    & 1006  & 83.5  \\
    Focal-B & 16.4  & 90    & 428   & 84.0  \\
    MPViT-B & 16.4  & 75    & 702   & 84.3  \\
    VMamba-B & 15.4  & 89    & 753   & 83.9  \\
    \rowcolor{gray!20}\textbf{SparX-Mamba-B} & 15.9  & 84    & 635   & 84.5  \\
    \bottomrule
    \end{tabular}%
}
  \caption{Comparison of throughput among representative CNN-, Transformer-, and Mamba-based models. The evaluation is conducted on a single NVIDIA H800 GPU using inputs of size 224$\times$224 and a batch size of 128.}
  \label{tab:speed-vit}%
\end{table}%

\begin{table*}[t]
  \centering
      \setlength{\tabcolsep}{1mm}
    \fontsize{9}{12}\selectfont 
    \resizebox{0.625\textwidth}{!}{
    \begin{tabular}{l|c|cc|ccc|ccc}
    \toprule
    \multirow{2}[4]{*}{Image Size/Batch Size} & EfficientVMamba & \multicolumn{2}{c|}{LocalVMamba} & \multicolumn{3}{c|}{VMamba} & \multicolumn{3}{c}{\textbf{SparX-Mamba}} \\
\cmidrule{2-10}          & Base  & Tiny  & Small & Tiny  & Small & base  & Tiny  & Small & base \\
    \midrule
    \textbf{224$^2$/128} &       &       &       &       &       &       &       &       &  \\
    Throughput (imgs/s) & 1943  & 597   & 1021  & 1613  & 1021  & 753   & 1366  & 871   & 635  \\
    Memory (MB) & 5308  & 10810  & 6745  & 6784  & 6745  & 8703  & 7066  & 7317  & 8807  \\
    Acc. (\%) & 81.8  & 82.7  & 83.6  & 82.5  & 83.6  & 83.9  & \textbf{83.5} & \textbf{84.2} & \textbf{84.5} \\
    \midrule
    \textbf{384$^2$/128} &       &       &       &       &       &       &       &       &  \\
    Throughput & 650   & 190   & 367   & 584   & 367   & 258   & 495   & 313   & 223  \\
    Memory & 13080  & 29494  & 17207 & 17246  & 17207 & 22483  & 18934  & 18603  & 21501 \\
    Acc. (\%) & -     & -     & 83.8  & 82.5  & 83.8  & 84.1  & \textbf{84.0} & \textbf{84.8} & \textbf{84.8} \\
    \midrule
    \textbf{512$^2$/128} &       &       &       &       &       &       &       &       &  \\
    Throughput & 383   & 106   & 220   & 342   & 220   & 151   & 298   & 194   & 138  \\
    Memory & 21108  & 52206  & 29445  & 29484  & 29445  & 38695  & 32562  & 31971  & 36929  \\
    Acc. (\%) & -     & -     & 82.9  & 81.1  & 82.9  & 83.3  & \textbf{82.9} & \textbf{84.1} & \textbf{83.9} \\
    \midrule
    \textbf{768$^2$/64} &       &       &       &       &       &       &       &       &  \\
    Throughput & 166   & 47    & 93    & 147   & 93    & 65    & 121   & 78    & 49  \\
    Memory & 23556  & 57508  & 32941  & 32980  & 32941  & 22467  & 36466  & 35569  & 21427  \\
    Acc. (\%) & -     & -     & 80.3  & 76.1  & 80.3  & 80.6  & \textbf{79.5} & \textbf{81.1} & \textbf{81.0} \\
    \midrule
    \textbf{1024$^2$/32} &       &       &       &       &       &       &       &       &  \\
    Throughput & 95    & 28    & 55    & 85    & 55    & 38    & 65    & 41    & 29  \\
    Memory & 21106  & 50830  & 29445  & 29226  & 29445  & 38695  & 32338  & 31859  & 36981  \\
    Acc. (\%) & -     & -     & 73.7  & 62.3  & 73.7  & 74.8  & \textbf{71.8} & \textbf{74.9} & \textbf{75.7} \\
    \bottomrule
    \end{tabular}%
   }
  \caption{Comparison of speed and GPU memory consumption among different Mamba-based hierarchical models on a single NVIDIA H800 GPU. Due to the unavailability of models that match our small and base sizes in EfficientVMamba, and the lack of a base model in LocalVMamba, we are unable to provide direct comparisons with these three models. Note that the complexity of EfficientVMamba-Base is on par with our tiny model.}
  \label{tab:speed_mamba}%
\end{table*}%

\subsection{Comparisons with More Advanced Models}
\label{appendix_compare}
To present a more comprehensive performance analysis, we compare our SparX-Mamba with highly optimized vision backbones in terms of image classification, semantic segmentation, and object detection. The methods compared include Conv2Former \cite{HouConv2Former}, InternImage \cite{wang2022internimage}, PeLK \cite{chen2024pelk}, CMT \cite{guo2022cmt}, MaxViT-T \cite{tu2022maxvit}, BiFormer \cite{zhu2023biformer}, MPViT \cite{lee2022mpvit}, NAT \cite{hassani2023neighborhood}, and CrossFormer++ \cite{wang2023crossformer++}. On the image classification task (Table~\ref{tab:in1k-p2}), the performance of our model is closer to state-of-the-art (SOTA) ViTs, CNNs, and their hybrid versions than other Mamba-based models. For example, SparX-Mamba-T is on par with InternImage-T~\cite{wang2022internimage} in terms of top-1 accuracy, and it only lags behind MaxViT-T~\cite{tu2022maxvit} by 0.2\%. In fact, our model outperforms most of those advanced CNN and ViT architectures, that is, SparX-Mamba-T surpasses CrossFormer++-S~\cite{wang2023crossformer++} and NAT-T~\cite{hassani2023neighborhood} by 0.3\% in accuracy. When the model is scaled up to larger sizes, similar phenomena can also be observed. More importantly, although SparX-Mamba may not perform best on image classification, it exhibits superior performance on dense prediction tasks with larger input resolutions. As shown in Tables \ref{tab:seg-p2} and \ref{tab:det-p2}, while SparX-Mamba-T and InternImage-T have comparable performance on image classification, SparX-Mamba-T surpasses InternImage-T by a large margin on both object detection and semantic segmentation. Although some highly optimized models have better performance than SparX-Mamba on image classification, SparX-Mamba ends up with better performance on dense prediction tasks. For example, CrossFormer++-L is superior to SparX-Mamba-B on image classification, but the latter clearly outperforms the former on semantic segmentation. Note that other Mamba-based models do not fully reflect this advantage. The superior performance of SparX-Mamba in dense prediction tasks can be potentially explained by its larger-than-usual effective receptive fields (ERFs), as visualized in Appendix \ref{sec:erf}. Furthermore, recent research \cite{yu2024mambaout} has shown that SSM can perform better on dense prediction tasks with high-resolution inputs, despite no performance advantage on the image classification task.
\subsection{Analytical Experiments}
\subsubsection{Impact of increasing resolutions:}
Following VMamba, we further assess the generalization capability of our model across different input resolutions. To be specific, we utilize models pre-trained on ImageNet-1K with input size of 224$\times$224 to perform inference on a range of image resolutions, from 384$\times$384 to 1024$\times$1024. As shown in Table~\ref{tab:resolutions}, our model exhibits the most stable performance as the input resolution increases, surpassing other competing models significantly. When the 384$\times$384 resolution is used, all other models experience performance degradation, whereas our SparX-Mamba shows a noticeable performance improvement, raising accuracy from 83.5\% to 84.0\%. Furthermore, as the resolution further increases, SparX-Mamba demonstrates the smallest performance drop. Additionally, it can be observed that as the resolution increases, the advantage of VMamba over CNN and Transformer models diminishes, whereas our model consistently maintains a significant performance advantage. For instance, at the 1024$\times$1024 input resolution, VMamba only achieves a marginal improvement of 0.4\% in comparison to Swin while our model outperforms Swin by nearly 10\%. These results demonstrate that our approach can better handle situations with a large number of input tokens.
\subsubsection{Speed comparisons among representative vision backbones:} 
\label{sec:appendix_speed}
We have conducted a comprehensive comparison of throughput (T) among representative vision backbones.
As shown in Table~\ref{tab:speed-vit}, recent Mamba-based models do not exhibit a speed advantage over classical ConvNeXt and Swin models. This is probably because deep learning frameworks have not been fully optimized to carry out SSM computations efficiently on GPUs, unlike matrix multiplications which have been highly optimized on GPUs. Meanwhile, the lower parallelization capacity of SSMs is another contributing factor in comparison to convolutions and self-attention. Despite these reasons, our network architecture still achieves a favorable trade-off between speed and accuracy in comparison to some advanced vision transformers. For example, SparX-Mamba-T outperforms MPViT-S in accuracy by notable 0.5\% top-1 accuracy while only experiencing a 10\% speed reduction. Our model also demonstrates significant advantages in both speed and performance when compared to Focal-Transformer. 
\par
On the other hand, compared with Mamba-based models, our SparX-Mamba also exhibits a better speed-accuracy trade-off. Specifically, when transitioning from VMamba-T to our SparX-Mamba-T, the top-1 accuracy improves significantly by 1\%, while the throughput only decreases by 243 imgs/s (from 1613 to 1370), accompanied by a reduction in model parameters. In contrast, when VMamba-T evolves into VMamba-S, the top-1 accuracy improves by 1.1\%, but the speed decreases substantially by 592 imgs/s (from 1613 imgs/s to 1021 imgs/s). Therefore, for a similar improvement of around 1\% in top-1 accuracy, SparX-Mamba-T demonstrates a significant advantage in throughput over VMamba-S (1370 vs. 1021), with only about half the model complexity. This clearly illustrates that our proposed architecture can provide a better speed-accuracy trade-off. Meanwhile, this advantage is also applicable to larger models. Specifically, our SparX-Mamba-S outperforms VMamba-B in terms of speed and model complexity, with a faster speed, lower model complexity, and better performance. Therefore, considering both speed and accuracy, the results clearly demonstrate that our SparX-Mamba has a significant speed-accuracy trade-off advantage over VMamba. Moreover, it is worth noting that EfficientVMamba's stage 3 and 4 are fully convolutional layers without any Mamba layers, making it unfair to compare computational efficiency with our method directly. Overall, our SparX-Mamba stands out among other transformer-based and Mamba-based models by striking an excellent balance between computational efficiency and performance.
\subsubsection{Speed comparisons among Mamba-based models on inputs with varying resolutions:} 
\label{sec:appendix_speed_mamba}
We conduct a comprehensive comparison of speed and GPU memory consumption among state-of-the-art Mamba-based hierarchical models on input images with different resolutions. As shown in Table \ref{tab:speed_mamba}, our model achieves the best trade-off between speed and performance in comparison to other Mamba-based models. It is worth mentioning that despite more connections among layers, our model does not significantly increase GPU memory consumption and latency in comparison to VMamba. This is because SparX is an efficient mechanism for creating cross-layer connections in Mamba-based models. Furthermore, an interesting phenomenon is that our small model exhibits a slight improvement over the base model when performing inference directly on some higher resolutions. The underlying reason may be that the model is trained on 224$\times$224 resolution, which limits its understanding of the knowledge contained in larger resolution images. The results suggest that our small model has learned more out-of-distribution knowledge during training, resulting in slightly better performance than the base model. However, when the model is trained directly on high-resolution images, the base model has a clear advantage over the small model, as evident in the performance of dense prediction tasks involving larger resolution inputs.
\subsubsection{Effective Receptive Field (ERF) analysis:}
\label{sec:erf}
\begin{figure}[!t]
    \centering
    \includegraphics[width=0.425\textwidth]{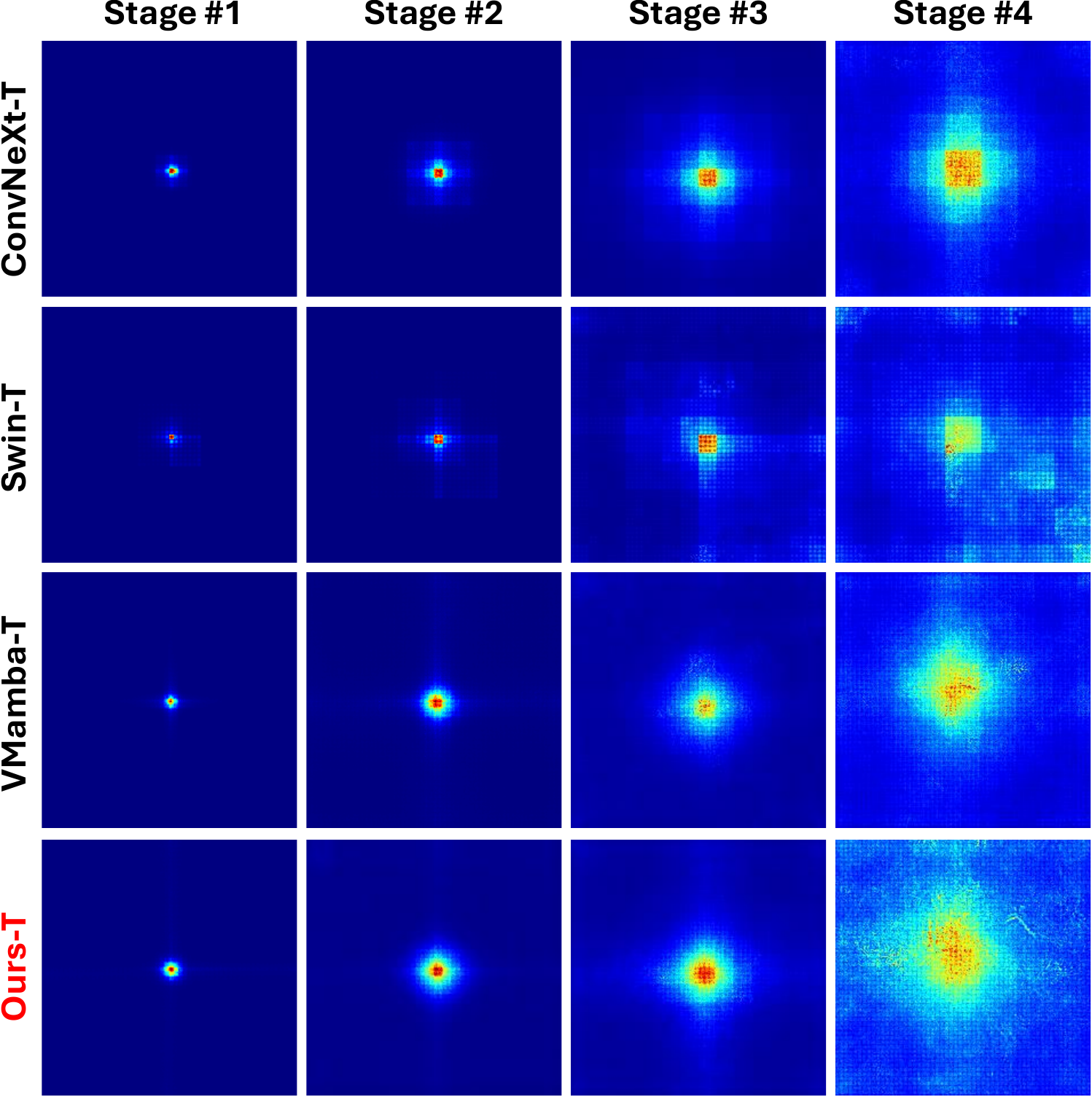}
    \caption{Visualization of the effective receptive fields of representative models.} 
    \label{fig:erf}
\end{figure}

To visually showcase the representation capacity of our SparX-Mamba, we visualize the ERF~\cite{luo2016understanding} of our model as well as other representative models, including ConvNeXt, Swin, and VMamba. The visualizations are generated using 200 randomly sampled images with 224$\times$224 resolution from ImageNet-1K. As shown in Figure~\ref{fig:erf}, it is evident that our model achieves the largest ERF across all stages of the network. In particular, the ERF of the final stage in SparX-Mamba nearly encompasses the entire input image, demonstrating its superior global context modeling capability.

\subsubsection{Centered Kernel Alignment (CKA) analysis:}
\label{sec:cka}
\begin{figure}[!t]
    \centering
    \includegraphics[width=0.475\textwidth]{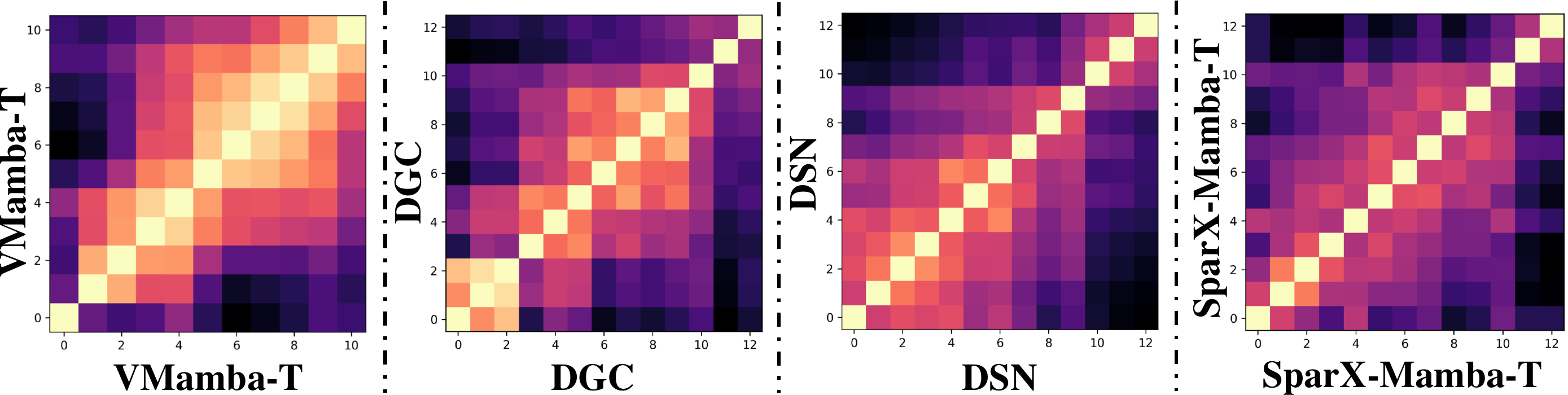}
    \caption{CKA analysis of model variants}
    \label{fig:cka}
\end{figure}

Since models with denser cross-layer connections (i.e., DGC and DSN models in Table \ref{tab:compare_densenet}) do not give rise to improved performance despite having more connections and higher computational complexity, we wish to discover the underlying reasons by analyzing layerwise features and calculating the similarity of learned feature patterns across layers using Centered Kernel Alignment (CKA)~\cite{kornblith2019similarity}. We chose four representative models: VMamba-T, the DGC and DSN models from Table~\ref{tab:compare_densenet}, and SparX-Mamba-T. In this analysis, features output from the SSM in each layer are used. As shown in Figure~\ref{fig:cka}, the patterns learned within each layer of VMamba-T are quite similar to those in nearby layers, a phenomenon also observed in classical ViT and CNN models~\cite{raghu2021vision}. It is evident that the layers of both DGC and DSN models can learn more unique patterns, which aids in developing more discriminative representations. However, when we sparsify cross-layer connections to obtain SparX-Mamba-T, the model learns even more diverse features across layers compared to other methods. Consequently, model performance is improved. According to this analysis, our biomimetic SparX promotes feature interaction and distillation by encouraging the model to learn more diverse representations. Although the DGC and DSN models promote feature diversity, our SparX model further boosts this diversity, which is the reason why it achieves better performance.

\subsection{Limitations}
\label{sec:limit}
Some highly optimized CNNs and ViTs show better performance than our method on the ImageNet-1K dataset as Mamba-based models are still in the early stages of exploration. However, it is worth noting that our SparseCC-Mamba model has already achieved performance close to many advanced CNNs and ViTs, and we have provided more comprehensive comparisons in Appendix \ref{appendix_compare}. On the other hand, compared to convolution and self-attention, SSMs have a more complex architecture. Existing deep learning frameworks only support limited optimization for SSMs, resulting in Mamba-based models not having a speed advantage. More details are given in Appendix \ref{sec:appendix_speed}. Finally, in addition to the three versions of our model provided in this work, we will explore the potential of using Mamba-based architectures for building large vision foundation models in the future.


\bibliography{aaai25}

\end{document}